\documentclass[runningheads]{llncs}
\usepackage[T1]{fontenc}
\usepackage{graphicx}
\usepackage{subcaption}
\usepackage{amsmath}
\usepackage{amsfonts}
\usepackage{tabularx}
\usepackage{makecell}
\usepackage{multirow}
\usepackage{rotating}
\usepackage{float}
\usepackage{longtable,verbatim}
\usepackage{xcolor}
\usepackage{hyperref}
\usepackage{pifont}
\usepackage{booktabs}
\usepackage{enumitem}
\usepackage{pifont}
\usepackage[misc]{ifsym}

\usepackage{booktabs}
\usepackage[table]{xcolor}
\usepackage{pgfplots}
\pgfplotsset{compat=1.18}
\usepackage{siunitx}
\definecolor{Dark}{HTML}{FFF0F7}
\definecolor{Light}{HTML}{E5F6FF}

\newcommand{\repeatthanks}{\textsuperscript{\thefootnote}}

\newcommand{\corr}{(\Letter)}

\newcolumntype{C}{>{\centering\arraybackslash}X}

\begin{document}

\title{TerraFlow: Multimodal, Multitemporal Representation Learning for Earth Observation}

\titlerunning{TerraFlow: Multimodal, Multitemporal Representation Learning for EO}


\author{Nazar Puriy\thanks{Equal contribution.}\inst{1,2} \and
Johannes Jakubik\repeatthanks\inst{1} \corr \and
Benedikt Blumenstiel\inst{1} \corr \and
Konrad Schindler\inst{2}
}

\authorrunning{N. Puriy, J. Jakubik, B. Blumenstiel, and K. Schindler}

\institute{IBM Research Europe, Zurich, Switzerland 
\and
ETH Zurich, Zurich, Switzerland
\email{\{johannes.jakubik1,benedikt.blumenstiel\}@ibm.com}
}

\maketitle              

\begin{abstract}
We propose TerraFlow, a novel approach to multimodal, multitemporal learning for Earth observation. TerraFlow builds on temporal training objectives that enable sequence-aware learning across space, time, and modality, while remaining robust to the variable-length inputs commonly encountered in real-world Earth observation data. Our experiments demonstrate superiority of TerraFlow over state-of-the-art foundation models for Earth observation across all temporal tasks of the GEO-Bench-2 benchmark. We additionally demonstrate that TerraFlow is able to make initial steps towards deep-learning based risk map prediction for natural disasters---a task on which other state-of-the-art foundation models frequently collapse. TerraFlow outperforms state-of-the-art foundation models by up to 50\% in F1 score and 24\% in Brier score.
\keywords{Earth observation \and Remote sensing \and Multitemporal \and Multimodality}
\end{abstract}

\section{Introduction}
\label{sec:introduction}
Earth observation (EO) is inherently multimodal and heterogeneous: the same location can be described by optical imagery, SAR, elevation, land-cover products, and other derived layers, each with different resolutions, noise characteristics, and acquisition constraints. At the same time, many high-value applications are fundamentally temporal as environmental processes evolve, disasters unfold, and observations are often irregular due to revisit patterns and atmospheric effects. This combination makes EO a challenging setting for building general-purpose vision models that transfer across regions, modalities, and tasks.
Recent EO foundation models have made substantial progress by leveraging large collections of unlabeled data and multimodal alignment. In particular, any-to-any generative pretraining has emerged as a strong paradigm for learning shared latent spaces across modalities~\cite{Jakubik2025TerraMind}, enabled by global-scale datasets with spatiotemporal aligned input modalities
\cite{Blumenstiel2025TerraMesh}. However, large-scale multimodal models often fall short in modeling multitemporal sequences at scale. 

\begin{figure}[h]
  \centering
  \includegraphics[width=\linewidth]{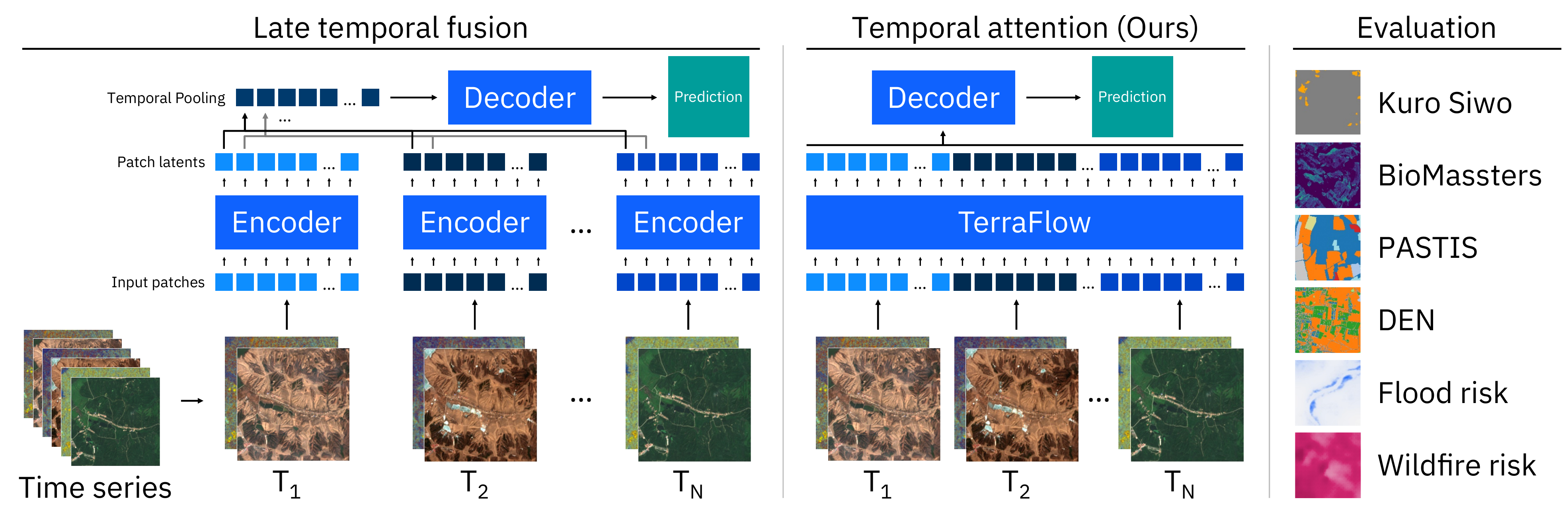}
  \caption{Image encoders can be applied to temporal tasks using a late fusion approach (left) while TerraFlow uses temporal attention for early fusion (middle). We benchmark TerraFlow on four temporal GEO-Bench-2 datasets and challenging disaster risk maps from ImpactMesh and Kuro Siwo (right).}
  \label{fig:terraflow_main_fig}
\end{figure}

We introduce TerraFlow to explicitly model temporal structure during pretraining, enabling sequence-level learning across space, time, and modality while remaining robust to sparse and variable-length inputs. In order to avoid the prohibitive cost of full retraining, we adopt a continuous pretraining strategy and initialize TerraFlow from the TerraMind foundation model~\cite{Jakubik2025TerraMind}. We then extend the resulting model with temporal objectives using large-scale multitemporal EO data. This design allows the model to preserve strong cross-modal representations while explicitly acquiring temporal learning. We evaluate TerraFlow under a rigorous and standardized benchmarking protocol across all temporal tasks of the GEO-Bench-2 benchmark~\cite{Simumba2025GeoBench2}, where it consistently outperforms both single-temporal and multi-temporal approaches. Beyond standard downstream evaluation, we further stress-test temporal generative pretraining in more challenging forecasting-oriented settings, including flood and wildfire risk map prediction~\cite{Bountos2024KuroSiwo,impactmesh2025}. In these scenarios, TerraFlow exhibits substantially larger gains over other foundation models, highlighting the importance of temporal pretraining for complex, high-impact EO applications. We summarize TerraFlow with the proposed temporal attention at the code at a high level in Fig.~\ref{fig:terraflow_main_fig}. 

Our main contributions are three-fold: (i) TerraFlow unites multimodal \emph{and} multitemporal generative pretraining in a single framework that strengthens temporal representations while supporting variable-length sequences; (ii) We propose Temporal Disjoint Sampling (TDS) as a novel strategy to natively merge the modeling of temporal sequences with masked modeling; and (iii) We demonstrate the superiority of TerraFlow over the state-of-the-art within an extensive evaluation on community-standard EO benchmarks with temporal tasks as well as in demanding forecasting-oriented settings. 

\section{Background}
\label{sec:relatedwork:remotesensingfoundationmodels}
\textbf{Remote Sensing Foundation Models.}
Geospatial foundation models (GeoFMs) adapt large-scale self-supervised pretraining to EO, leveraging abundant unlabeled data to improve downstream performance and generalization under limited labeled supervision. Masked image modeling (MIM) has been particularly effective: SatMAE extends MAE-style objectives to multispectral and short temporal stacks~\cite{Cong2022SatMAE}, while Scale-MAE addresses resolution/scale shifts common in EO imagery~\cite{Reed2023ScaleMAE}. Moving beyond single-sensor pretraining, CROMA combines contrastive radar--optical alignment with masked reconstruction~\cite{Fuller2023CROMA}, and msGFM studies multi-sensor representations via cross-sensor reconstruction~\cite{Han2024msGFM}. Prithvi-style models scale multitemporal pretraining over large archives, with Prithvi-EO-2.0 adding stronger multitemporal coverage and metadata conditioning~\cite{Jakubik2023Prithvi,Szwarcman2024PrithviEO2}. For any-to-any generative multimodality, TerraMind introduces a token prediction objective over many modalities~\cite{Jakubik2025TerraMind}. \\

\noindent\textbf{Temporal Modeling.}
Temporal EO learning has been approached with recurrent models (e.g., based on LSTM architectures) that propagate information over time~\cite{Hochreiter1997LSTM,Shi2015ConvLSTM}, 3D CNNs that treat time as an additional axis~\cite{Tran2015C3D}, and Transformers that use attention to capture long-range dependencies under irregular sampling~\cite{tarasiou2023vits}. In modern EO pretraining, time is increasingly injected explicitly (temporal embeddings/masking/objectives), as in SatMAE and Prithvi-EO-2.0, and in spatiotemporal foundation-model designs such as Galileo and TiMo~\cite{Cong2022SatMAE,Qin2025TiMo,Szwarcman2024PrithviEO2,Tseng2025Galileo}. A key practical constraint is robustness to sparse, variable-length sequences caused by clouds and acquisition gaps.\\

\noindent\textbf{AI for Disaster Prediction.}
Most EO-based natural disaster prediction aim to \emph{map} the target area after (or during) an event by supervised segmentation of SAR and/or optical imagery, supported by datasets like Sen1Floods11~\cite{Bonafilia2020Sen1Floods11} and large multitemporal SAR benchmarks like Kuro Siwo~\cite{Bountos2024KuroSiwo}. These efforts enable rapid impact assessment, but they do not directly address the predictive estimation of \emph{future} flood risk from pre-event context. Operational \emph{forecasting} primarily relies on physics-based hydrological pipelines driven by meteorological forcings, exemplified by LISFLOOD-style river-basin modeling and global early-warning systems such as GloFAS~\cite{Alfieri2013GloFAS,VanDerKnijff2010LISFLOOD}. 
While effective at large spatial scales, these systems exhibit limitations in resolving high-resolution inundation extents and remain sensitive to uncertainties in forcings and local conditions. Recent large-scale multitemporal flood datasets—--notably Kuro Siwo~\cite{Bountos2024KuroSiwo} for flood events and ImpactMesh~\cite{impactmesh2025} for both flood and wildfire events—--provide broad event coverage and aligned EO inputs with annotations. Together with large-scale temporal pretraining, these resources make it increasingly feasible to explore learning-based forecasting of future disaster impacts from pre-event EO context, opening the door to approaches that complement and integrate with traditional physical modeling systems.

\section{Methodology}
\label{sec:methodology}

In the following, we introduce our methodology by describing the underlying architecture, the temporal encoding strategy, and the masking strategy.\\

\begin{figure}[h]
  \centering
  \includegraphics[width=\linewidth]{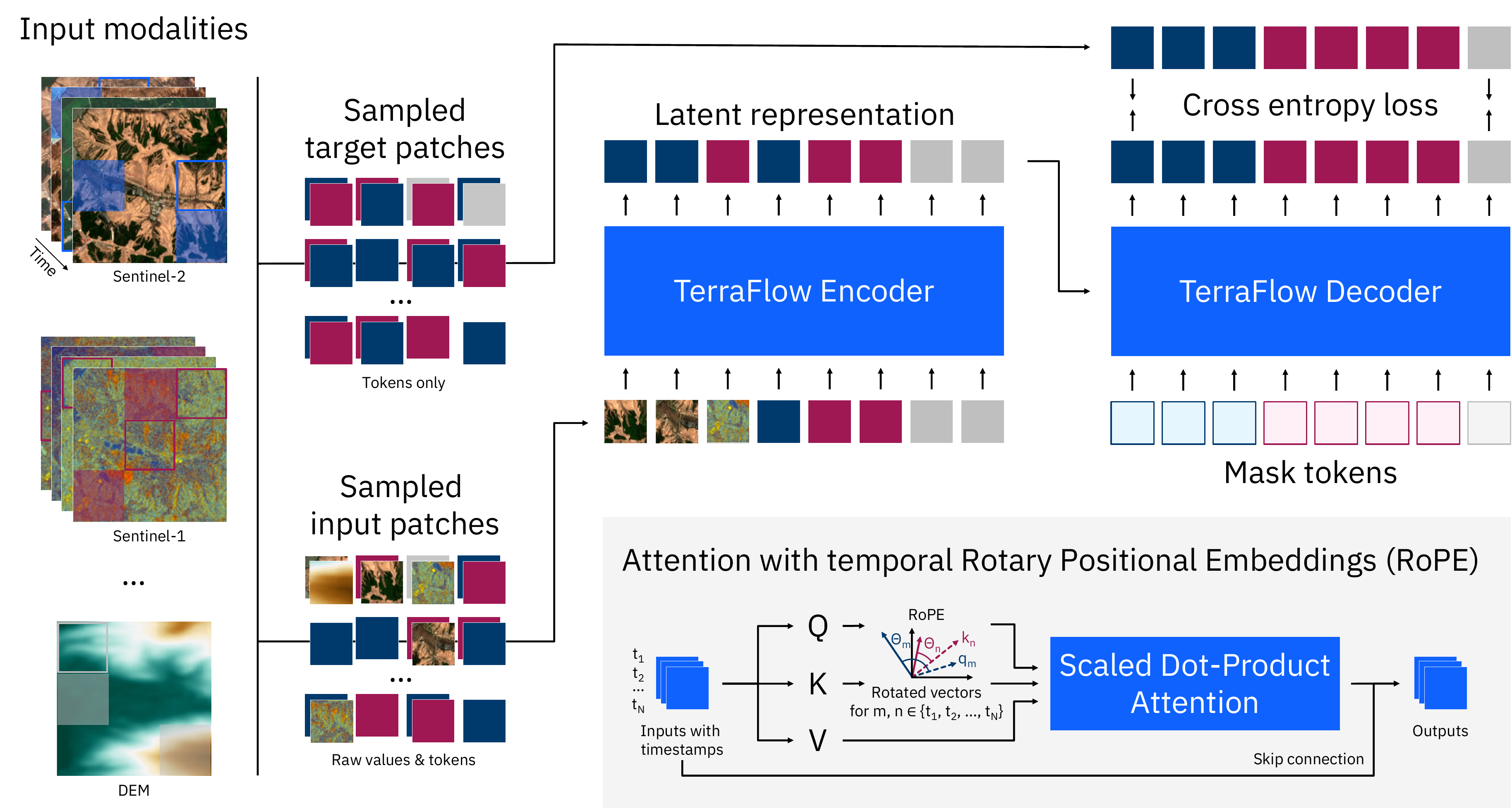}
  \caption{\textbf{TerraFlow pretraining.} Input and target patches are sampled from multiple modalities and timestamps, represented by either raw pixel values or tokens. TerraFlow uses a standard transformer encoder-decoder architecture, trained with cross entropy loss. The attention blocks apply RoPE to encode the relative temporal offset between timestamps in the queries and keys.}
  \label{fig:terraflow_overview}
\end{figure}

\noindent\textbf{Architecture Overview.}
TerraFlow follows a Transformer encoder--decoder design for masked multimodal prediction~\cite{Mizrahi2023FourM}. We generally adopt the pretraining strategy of TerraMind~\cite{Jakubik2025TerraMind} and extend it for large-scale multitemporal learning, by continuously pretraining the encoder-decoder architecture. I.e., we maintain the same approach to multimodality, leveraging dual-scale representations in the form of pixels and quantized tokens that we then feed to the model encoder across diverse modalities. The task of the model is to predict target tokens similar to masked reconstruction across modalities but formulated as a classification task over all tokens. For details on the general approach to multimodality in TerraMind, we refer to Jakubik et al.~\cite{Jakubik2025TerraMind}. In contrast to TerraMind, TerraFlow introduces large-scale temporal learning for multimodal models. Therefore, we introduce the following temporal design: (i)~We implement rotary positional encoding across the temporal dimension as part of the \textbf{Temporal Encoding Strategy} and (ii)~introduce \textbf{Temporal Disjoint Sampling} as a novel sampling strategy for multimodal masked reconstruction, where the temporal input sequence is randomly divided into temporally disjoint subsets. In this way, the input sequence to TerraFlow contains tokens from multiple modalities and multiple timestamps jointly, enabling {early fusion} of temporal information: attention can capture cross-temporal, cross-modal interactions within the forward passes. We summarize this approach at a high level in Fig.~\ref{fig:terraflow_overview}.\\

\noindent\textbf{Temporal Encoding Strategy.}
For temporal encoding, we leverage rotary positional encoding (RoPE)~\cite{su2024roformer}. Our rationale is as follows: First, RoPE enables explicit modeling of relative temporal positions, which is particularly important for Earth observation time series, where absolute timestamps are often implicitly available from context and observations are irregularly sampled. Second, rotary encoding affects only the attention interactions between tokens, which significantly benefits continual pretraining, as the underlying embedding space remains largely unchanged. Third, for a single timestamp, i.e., when $n=m$, the relative rotation matrix $R_{\Theta,(n-m)}^d$ becomes identity, meaning the attention calculation remains unchanged compared to TerraMind, so we can leverage spatial pretraining, and all previously learned interactions are processed in the same way. We implement temporal RoPE for TerraFlow by applying a time-dependent rotation to query and key projections. That means, for a token embedding $x \in \mathbb{R}^d$ observed at time $m$, we generate a function $f$ that rotates projections as follows

\begin{equation}
\label{eq:rotatoryattention}
f_{\{q,k\}}(x_m,m) = R_{\Theta,m}^d W_{\{q,k\}} x_m,
\end{equation}

\noindent where $R_{\Theta,m}^d$ \eqref{eq:rotationmatrix} corresponds to rotations in 2D planes

\begin{equation}
\label{eq:rotationmatrix}
R_{\Theta,m}^d
=
\mathrm{diag}\!\left(
\left\{
\begin{bmatrix}
\cos(m\theta_i) & -\sin(m\theta_i) \\
\sin(m\theta_i) & \cos(m\theta_i)
\end{bmatrix}
\right\}_{i=0}^{\frac{d}{2}-1}
\right),
\end{equation}

\noindent where we use predefined angular frequencies $\theta_i = 10000^{-2i/d}$ with $i=0,\dots, \break\tfrac{d}{2}-1$. As mentioned, a resulting key property is that the attention then depends only on relative temporal offsets. Specifically, for tokens at times $m$ and $n$, the encoding depends only on their relative distance $n - m$ as follows
\begin{equation}
\label{eq:relattention}
\langle q_m, k_n \rangle
=
x_m^\top W_q^\top R_{\Theta,(n-m)}^d W_k x_n.
\end{equation}\\

\noindent We use the number of days since January 1, 2000 as the value for $m$. We drop hour-level information, as it does not provide additional useful signal for the tasks considered, and models can typically infer such temporal cues directly from the images.\\

\noindent\textbf{Masking Strategy.}
\label{sec:methodology:maskingstrategy}
A core goal of TerraFlow is to maintain the strong cross-modal and spatial correlations learned by large-scale multimodal pretraining, while additionally learning temporal structure. We hypothesize that, when continuing pretraining from a non-temporal checkpoint, naive masking (random tokens across all timestamps) can lead to a regime where the model largely solves the objective with single-timestep shortcuts and remains near its previous optimum. This would naturally limit temporal learning. Therefore, to explicitly enforce temporal learning, we introduce a new sampling regime. Before sampling tokens, we first randomly partition the available timestamps into disjoint, non-empty input and target sets. This is done by uniformly selecting between one and $N-1$ timestamps as input and using the remaining timestamps as output, where $N=4$ for the used dataset. Within each selected timestamp, tokens are then sampled following Dirichlet distributions. This design ensures that the decoder must predict tokens using timestamps different from the ones provided as input, encouraging the model to exploit temporal cues (e.g., change patterns, persistence, and dynamics), rather than relying solely on spatial correlations at a single timestamp. We call this approach \textbf{Temporal Disjoint Sampling (TDS)}, in opposition to \textbf{Random Sampling (RS)}. We pretrain the model under both regimes and report experimental results to evaluate our hypothesis.

\section{Experimental Setting}\label{appendix:exp_setting}

\noindent In the following, we detail the pretraining setup. For reasons of space, we provide additional details on the downstream adaptation setup in Section~\ref{appendix:exp_setting} of the supplementary material.\\

\noindent\textbf{Pretraining Dataset.}
We select the SSL4EO-S12 v1.1 dataset~\cite{blumenstiel2025ssl4eo,wang2023ssl4eo} for continual pretraining. SSL4EO-S12 represents the temporal subset of TerraMesh~\cite{Blumenstiel2025TerraMesh}, which contains Sentinel-1 GRD (SAR), Sentinel-2 L1C, L2A and RGB (optical), DEM (elevation), and land-cover modalities around the 10,000 most populated cities. Image chips have a size of 264$\times$264 pixels at 10m ground sampling distance (GSD). Temporal data is available for Sentinel-1, Sentinel-2, and land-cover, with four timestamps between 2019 and 2021 per geo-location corresponding to one sample per season. For the DEM, the dataset includes a single timestamp. Timestamps have been temporally aligned based on Sentinel-2 acquisitions. It is important to emphasize that TerraFlow is trained exclusively on SSL4EO-S12, a strict subset of the data used for TerraMind pretraining. The model is not exposed to any new, additional data. Hence, we can directly measure the effect of our temporal learning strategy.\\

\noindent\textbf{Pretraining Setup.}
We pretrain four variants of TerraFlow across two model sizes and the two previously introduced temporal sampling strategies. Model sizes are 5.5M for the \emph{Tiny} encoder and 85.5M for the \emph{Base} encoder. For each model size, we consider two temporal masking regimes TDS and RS, as described in Section~\ref{sec:methodology:maskingstrategy}. All models are initialized from publicly available TerraMind weights~\cite{Jakubik2025TerraMind}. Training is performed with the AdamW optimizer and a cosine learning-rate scheduler, using a maximum learning rate of $1.6 \times 10^{-4}$ for Base models and $6.4 \times 10^{-4}$ for Tiny models. All experiments are conducted in bfloat16 precision. Generally, we train all four variants on the same set of samples from the SSL4EO-S12 dataset resulting in a token budget of approximately 75 billion tokens per variant. 
Continual pretraining was carried out on four NVIDIA A100-80GB GPUs for eight and 14 days for Tiny and Base, resp.\\

\noindent\textbf{Downstream Datasets and Tasks.}
We strictly follow the official GEO-Bench-2 evaluation protocol and baseline configuration~\cite{Simumba2025GeoBench2}. For each model--dataset pair, hyperparameters are selected with a fixed search space and a maximum budget of 16 trials, after which the configuration with the lowest validation loss is re-trained and evaluated five times with different random seeds. The learning rate is selected from 
$10^{-6}$ to $10^{-3}$, and the batch size from 8, 16, and 32. 
Inputs are standardized with per-band Z-score normalization, using statistics computed for the training split; for regression datasets, the same normalization strategy is applied to target values. During training, we apply standard horizontal and vertical flips consistently across all modalities and timestamps within a sample.
We evaluate on the multi-temporal subset of settings included in GEO-Bench-2, spanning three semantic segmentation and one regression task with varying temporal densities and modalities. Relevant missing information on the number of timestamps used per dataset within GEO-Bench-2 was kindly provided by the paper authors and is made public within the dataset overview below:

\begin{itemize}
    \item \textbf{Kuro Siwo} targets multitemporal flood mapping using Sentinel-1 SAR data, optionally complemented by DEM elevation maps~\cite{Bountos2024KuroSiwo}. The task is formulated as semantic segmentation of permanent water, flooded and non-flooded areas. Each sample provides $T{=}3$ timestamps.
    \item \textbf{PASTIS} focuses on multitemporal crop-type segmentation from Sentinel-2 imagery, with optional Sentinel-1 inputs. While the original dataset contains long sequences (33–61 timestamps) ~\cite{Garnot2021PASTIS}, we follow the GEO-Bench-2 protocol and subsample $T{=}7$ timestamps for comparative evaluation.
    \item \textbf{BioMassters} addresses pixel-wise regression of forest above-ground biomass using multimodal Sentinel-1 and Sentinel-2 time series, supervised by LiDAR-derived biomass estimates~\cite{Nascetti2023BioMassters}. The dataset provides up to 12 monthly observations; we subsample them to $T{=}7$ for comparability with GEO-Bench-2.
    \item \textbf{DynamicEarthNet} is a land-use and land-cover semantic change segmentation dataset based on dense Planet Lab imagery with seven classes~\cite{Toker2022DynamicEarthNet}. We use the weekly temporal setting, resulting in $T{=}6$ timestamps for the PlanetScope modality and a static image for S2.
\end{itemize}

\noindent Beyond GEO-Bench-2, we evaluate TerraFlow in a more demanding, forecasting-oriented setting where it is strictly required to leverage temporal structure, namely \textbf{prediction of disaster risk maps}. While EO models are frequently used for post-event mapping, predicting the potential spatial footprint of an event from pre-event context is another relevant, but more challenging task. We also explore this additional, novel setting, since existing post-event benchmarks have become increasingly saturated, limiting the expressiveness of the benchmarks.
Our objective is not to claim state-of-the-art performance in disaster forecasting, let alone to present definitive risk maps. Rather, we use the task as an exploratory testbed to assess whether TerraFlow possesses sufficient representational capacity to engage with such complex, uncertainty-driven problems. 
We also see this set of tasks as a forward-looking direction that highlights open questions and opportunities for future multitemporal EO models.\\

\noindent We explore the creation of risk maps for floods and wildfires, using Kuro Siwo~\cite{Bountos2024KuroSiwo} and ImpactMesh~\cite{impactmesh2025}. For Kuro Siwo, flood events are defined as transient surface water expansions observable from SAR imagery, encompassing both high-impact disasters and large-scale inundation phenomena. The train, validation, and test splits contain samples drawn from strictly different events. ImpactMesh is a large-scale multimodal, multitemporal dataset for floods and fires with four timesteps per event (two \emph{pre} and two \emph{post} images), which integrates Sentinel-1 RTC, Sentinel-2 L2A and the DEM. The train, validation, and test splits are defined over different areas of interest and are strictly non-overlapping. Approximately one third of the validation and test sets correspond to entirely separate events. 
To adapt the data for the risk prediction task, we only input \emph{pre-event} images and train models to predict a pixel-wise probability map of the affected area, which can be interpreted as spatially explicit risk maps. For Kuro Siwo, due to the scarcity of flood events and the resulting class imbalance, we restrict the dataset to scenes in which a flood is present. Moreover, we found implausible outlier values in the DEM modality of several samples, which we filter out. 
\\

\begin{table}[tb]
\centering
\caption{Model comparison on GEO-Bench-2 temporal datasets. The top section reports publicly available GEO-Bench-2 results~\cite{Simumba2025GeoBench2} that could be reproduced. PASTIS and DynamicEarthNet (DEN) values are omitted for baselines due to discrepancies during reproducibility checks; see supplementary material~\ref{app:geobench2}. BioMassters reports \((1 - \mathrm{std}) \cdot \mathrm{RMSE}\) [\%] as proposed in GEO-Bench-2.}

\begin{tabularx}{\textwidth}{lcCCCC}
\toprule
 & & Kuro Siwo  & BioMassters & PASTIS & DEN  \\
Model & Params & mIoU [\%] ↑ & custom ↑ & mIoU [\%] ↑ & mIoU [\%] ↑ \\

\midrule
Clay-V1 ViT-B~\cite{ClayFoundationModel} & 100M 
& 68.6 $\pm$ 0.3 & 95.8 $\pm$ 0.0 & -- & -- \\
TerraMind 1.0 Base~\cite{Jakubik2025TerraMind} & 100M 
& 67.9 $\pm$ 0.3 & 95.5 $\pm$ 0.0 & -- & -- \\
DOFA ViT~\cite{Xiong2024DOFA} & 300M 
& 69.7 $\pm$ 0.2 & 95.6 $\pm$ 0.0 & -- & -- \\
DINOv3-ViT-L-SAT~\cite{Simeoni2025DINOv3} & 300M 
& 69.0 $\pm$ 0.2 & 95.5 $\pm$ 0.0 & -- & -- \\
TerraMind 1.0 Large~\cite{Jakubik2025TerraMind} & 300M 
& 68.6 $\pm$ 0.2 & \textbf{96.0 $\pm$ 0.0} & -- & -- \\
ConvNeXt-XL-IN~\cite{Liu2022ConvNeXt} & 390M 
& 68.6 $\pm$ 0.2 & 95.6 $\pm$ 0.0 & -- & -- \\
Prithvi-EO-2.0-TL~\cite{Szwarcman2024PrithviEO2} & 600M 
& 67.5 $\pm$ 0.1 & 95.6 $\pm$ 0.0 & -- & -- \\
\midrule
TerraMind 1.0 Tiny & 8M 
& $67.6 \pm 0.5$ & $95.8 \pm 0.0$ & $50.5 \pm 0.5$ & $30.2 \pm 1.0$ \\
TerraFlow-Tiny-RS & 8M 
& $67.7 \pm 0.7$ & $95.9 \pm 0.0$ & $54.0 \pm 0.3$ & $35.4 \pm 0.8$ \\
TerraFlow-Tiny-TDS & 8M 
& $66.5 \pm 0.4$ & $95.9 \pm 0.0$ & $51.4 \pm 0.6$ & $33.5 \pm 0.9$ \\
\midrule
TerraMind 1.0 Base & 100M 
& $67.5 \pm 0.7$ & $95.9 \pm 0.0$ & $54.3 \pm 0.5$ & \underline{$35.8 \pm 1.1$} \\
TerraFlow-Base-RS & 100M 
& \underline{$70.0 \pm 1.3$} & \boldmath{$96.0 \pm 0.0$} & \boldmath{$58.0 \pm 0.4$} & \underline{$35.8 \pm 1.6$} \\
TerraFlow-Base-TDS & 100M 
& \boldmath{$70.1 \pm 0.7$} & \boldmath{$96.0 \pm 0.0$} & \underline{$56.9 \pm 0.4$} & \boldmath{$38.5 \pm 1.0$} \\
\bottomrule
\end{tabularx}

\label{tab:geobench_results}
\end{table}

\section{Experiments}
\label{sec:results}

This section presents results across GEO-Bench-2 and disaster prediction tasks, demonstrating that our approach consistently outperforms existing baselines, including TerraMind, across a wide range of temporal benchmarks. \\

\noindent\textbf{{GEO-Bench-2.}}
In Table~\ref{tab:geobench_results}, we report the performance of our models together with the publicly available results from the GEO-Bench-2 leaderboard. These baselines include a diverse set of strong foundation models, spanning different modeling paradigms. 
Among these baselines, TerraMind achieves the strongest overall performance on the GEO-Bench-2 temporal benchmarks. For this reason, we re-run TerraMind under the same evaluation protocol in order to (i)~reproduce the GEO-Bench-2 result where possible and (ii)~enable a direct and controlled comparison with our proposed TerraFlow models.
Under a controlled evaluation setup, TerraFlow consistently outperforms TerraMind at comparable model scales, indicating that the gains stem from improved temporal modeling rather than increased capacity. At 100M parameters, TerraFlow-Base TDS and RS consistently outperform existing models, improving over TerraMind on Kuro Siwo  (70.1 vs.\ 67.5 mIoU), PASTIS (58.0 vs.\ 54.3 mIoU), and DynamicEarthNet (38.5 vs.\ 35.8 mIoU). On BioMassters, TerraFlow matches the best reported score (96.0), indicating that the added temporal modeling capabilities do not compromise per-frame predictive accuracy. Notably, the 8M TerraFlow-Tiny-RS models perform comparably to the 100M TerraMind baseline across multiple datasets, highlighting the parameter efficiency of the proposed approach and the effectiveness of its temporal representation learning.\\

\begin{table}[!h]
\centering
\setlength{\tabcolsep}{4pt}
\caption{Comparison of Kuro Siwo  flood risk prediction performance across models. The top section reports models provided by Kuro Siwo~\cite{Bountos2024KuroSiwo} followed by Tiny and Base variants of TerraMind and TerraFlow. 
} 
\begin{tabular}{lcccccc}
\toprule
Model & PT & TA & Loss & F1$_{Flood}$ [\%] ↑ & Brier [\%] ↓ \\
\midrule
ResNet18 & \ding{55} & \ding{55} & CE & 23.6 ± 0.3 & 14.8 ± 0.8 \\
ResNet50 & \ding{55} & \ding{55} & CE & 23.0 ± 0.9 & 15.5 ± 0.8 \\
ResNet101 & \ding{55} & \ding{55} & CE & 22.6 ± 1.5 & 15.3 ± 1.7 \\
\midrule
ViT-Tiny & \ding{55} & \ding{51} & CE & 18.9 ± 0.0 & 11.6 ± 0.3 \\
TerraMind 1.0 Tiny & \ding{51} & \ding{55} & CE & 23.2 ± 0.1 & 12.1 ± 0.9 \\
TerraMind 1.0 Tiny & \ding{51} & \ding{51} & CE & 30.0 ± 0.7 & 11.0 ± 0.1 \\
TerraFlow-Tiny-RS & \ding{51} & \ding{51} & CE & 31.4 ± 0.6 & 11.3 ± 0.4 \\
TerraFlow-Tiny-TDS & \ding{51} & \ding{51} & CE & 26.8 ± 0.2 & 14.1 ± 0.2 \\
\midrule
ViT-Base & \ding{55} & \ding{51} & CE & 16.5 ± 0.2 & 12.2 ± 0.5 \\
TerraMind 1.0 Base & \ding{51} & \ding{55} & CE & 25.0 ± 1.0 & 16.3 ± 0.7 \\
TerraMind 1.0 Base & \ding{51} & \ding{51} & CE & 29.1 ± 0.4 & 10.2 ± 0.1 \\
TerraFlow-Base-RS & \ding{51} & \ding{51} & CE & \textbf{38.0 ± 1.0} & 9.7 ± 0.2 \\
TerraFlow-Base-RS & \ding{51} & \ding{51} & MSE & 35.3 ± 2.1 & \textbf{9.1 ± 0.3} \\
TerraFlow-Base-TDS & \ding{51} & \ding{51} & CE & \underline{36.0 ± 0.3} & 11.1 ± 0.1 \\
TerraFlow-Base-TDS & \ding{51} & \ding{51} & MSE & 35.2 ± 1.1 & \underline{9.5 ± 0.0} \\
\bottomrule
\end{tabular}
\label{tab:kurosiwoflood_comparison}
\end{table}

\subsection{Spatial Risk Map Prediction}

\subsubsection{Kuro Siwo.}
In Table~\ref{tab:kurosiwoflood_comparison}, we report the performances of TerraFlow and several recent baselines. Overall, TerraFlow-Base-RS outperforms other models, and its variant trained with MSE-objective achieves the best Brier score. TerraFlow-Base-TDS ranks second overall. Notably, TerraFlow-Tiny-RS outperforms the TerraMind-Base model. We additionally ablate the pre-training by also fine-tuning from randomly initialized weights (PT \ding{55}), which results in generally low performances. Initializing the model with TerraMind weights leads to a clear improvement. Adding temporal attention further boosts performance, and initializing with TerraFlow pretrained weights yields the best overall results. These findings indicate that both the model architecture and the pretrained weights play important roles. For further, detailed results of risk map prediction on additional datasets, as well as qualitative results, see the supplementary material. 

\begin{figure}[!ht]
    \centering
    \includegraphics[width=0.8\linewidth]{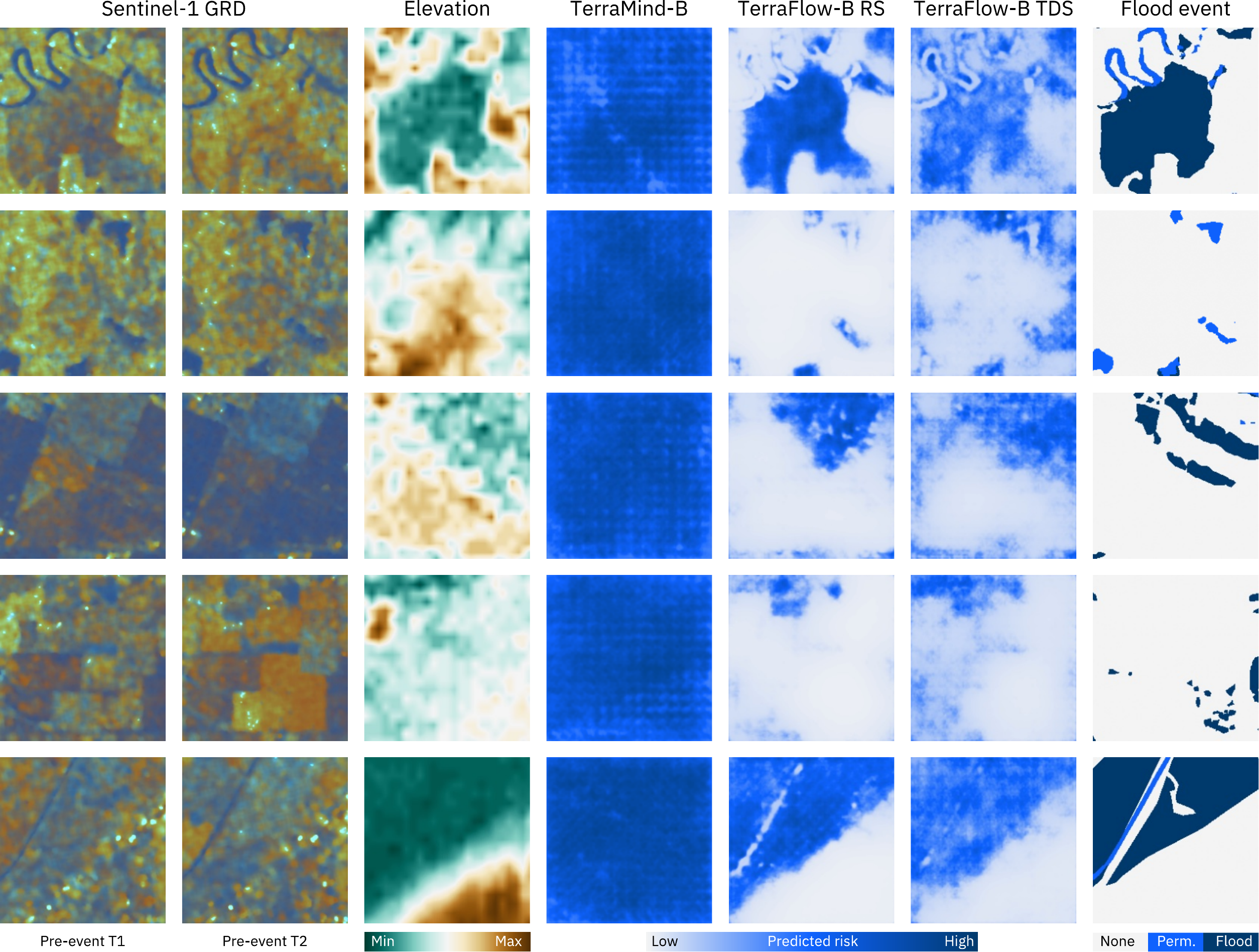}
    \caption{Qualitative comparison on the Kuro Siwo flood risk prediction task. The first columns display two pre-event Sentinel-1 images with VV-VH-VV/VH pseudo coloring, along with the DEM. TerraFlow predicts logical risk maps and understands permanent water bodies while image-level models like TerraMind completely fail to accurately predict risk. 
    }
    \label{fig:kurosiwo_comparison}
\end{figure}

A qualitative comparison of the spatial risk maps predicted by TerraMind and TerraFlow is shown in Fig.~\ref{fig:kurosiwo_comparison}. Generally, the results with TerraMind offer little visually meaningful information. In contrast, and in line with our expectation, spatial risk maps generated from TerraFlow avoid failure modes such as flood risk in permanent water bodies, and are generally better aligned with the actual flood extent in the unseen post-event image.

\subsubsection{ImpactMesh.}
In Table~\ref{tab:impactmesh_comparison}, we report similar findings for ImpactMesh, further showcasing the importance of both model architecture and pretrained weights. TerraFlow-Base-TDS outperforms all baselines on both flood and fire prediction. For floods, the MSE variant achieves the best Brier score, whereas for fire prediction, training with cross-entropy yields better Brier scores. We again observe that models initialized with random weights perform poorly, while initialization with general-purpose TerraMind weights leads to substantially better performance, which further improves when using the newly proposed approach. Notably, our pretrained TerraFlow-Tiny-TDS outperforms TerraFlow-Base initialized with TerraMind weights, which highlights the impact of effective pretraining beyond architectural scale. Like for Kuro Siwo, we show qualitative flood mapping results in the supplementary material, 
and fire risk maps in Fig.~\ref{fig:impactfires_comparison}. 
For flood risk, the qualitative analysis confirms our observations from the Kuro Siwo tests. 
In contrast, wildfire predictions show weak spatial differentiation across all models, with large portions of images frequently predicted as uniformly high risk. As a result, although quantitative metrics appear higher––—partly due to a stronger random baseline––—the qualitative wildfire maps are substantially worse. This likely reflects that wildfire risk depends not only on vegetation patterns visible in satellite imagery but are strongly influenced also by weather and human factors that are not directly observable.


\begin{table}[!h]
\centering
\caption{
Performance comparison on ImpactMesh for flood and fire risk prediction, highlighting the impact of pretrained weights~(PT), temporal attention~(TA), and the training loss.}
\label{tab:impactmesh_comparison}
\begin{tabularx}{\textwidth}{lcccCCCC}
\toprule
 & & & & \multicolumn{2}{c}{ImpactMesh-Flood} & \multicolumn{2}{c}{ImpactMesh-Fire} \\
Model & PT & TA & Loss & F1$_{Fl.}$ [\%] ↑ & Brier [\%] ↓ & F1$_{Fi.}$ [\%] ↑ & Brier [\%] ↓ \\
\midrule
ViT-Tiny & \ding{55} & \ding{51} & CE & 21.5 ± 2.9 & 5.8 ± 0.6 & 55.9 ± 0.8 & 21.1 ± 1.5 \\
TerraMind 1.0 Tiny & \ding{51} & \ding{55} & CE & 32.9 ± 0.8 & 6.2 ± 0.2 & 54.6 ± 1.0 & 20.9 ± 1.5 \\
TerraMind 1.0 Tiny & \ding{51} & \ding{51} & CE & 33.2 ± 1.9 & 5.2 ± 0.2 & 56.0 ± 0.5 & 20.3 ± 0.4 \\
TerraFlow-Tiny-RS & \ding{51} & \ding{51} & CE & 36.0 ± 1.1 & 5.0 ± 0.1 & 56.0 ± 1.7 & 19.4 ± 0.3 \\
TerraFlow-Tiny-TDS & \ding{51} & \ding{51} & CE & 38.1 ± 0.6 & 5.0 ± 0.1 & 56.3 ± 0.9 & \underline{19.2 ± 0.8} \\
\midrule
ViT-Base & \ding{55} & \ding{51} & CE & 24.0 ± 1.1 & 5.3 ± 0.2 & 55.8 ± 0.3 & 20.5 ± 0.7 \\
TerraMind 1.0 Base & \ding{51} & \ding{55} & CE & 36.3 ± 1.2 & 5.3 ± 0.3 & 54.8 ± 0.9 & 19.5 ± 0.6 \\
TerraMind 1.0 Base & \ding{51} & \ding{51} & CE & 36.9 ± 1.2 & 5.2 ± 0.1 & 56.2 ± 0.6 & 19.4 ± 0.6 \\
TerraFlow-Base-RS & \ding{51} & \ding{51} & CE & 38.2 ± 0.9 & 4.9 ± 0.2 & \underline{56.5 ± 0.6} & \underline{19.2 ± 0.6} \\
TerraFlow-Base-RS & \ding{51} & \ding{51} & MSE & 37.4 ± 1.2 & \textbf{4.8 ± 0.1} & 55.7 ± 0.6 & 19.9 ± 0.9 \\
TerraFlow-Base-TDS & \ding{51} & \ding{51} & CE & \underline{39.7 ± 0.4} & 4.9 ± 0.0 & \textbf{56.6 ± 0.3} & \textbf{18.6 ± 0.4} \\
TerraFlow-Base-TDS & \ding{51} & \ding{51} & MSE & \textbf{39.8 ± 1.3} & \textbf{4.8 ± 0.1} & 56.3 ± 0.4 & 20.1 ± 1.1 \\
\bottomrule
\end{tabularx}
\end{table}

\begin{figure}[tbh]
    \centering
    \includegraphics[width=\linewidth]{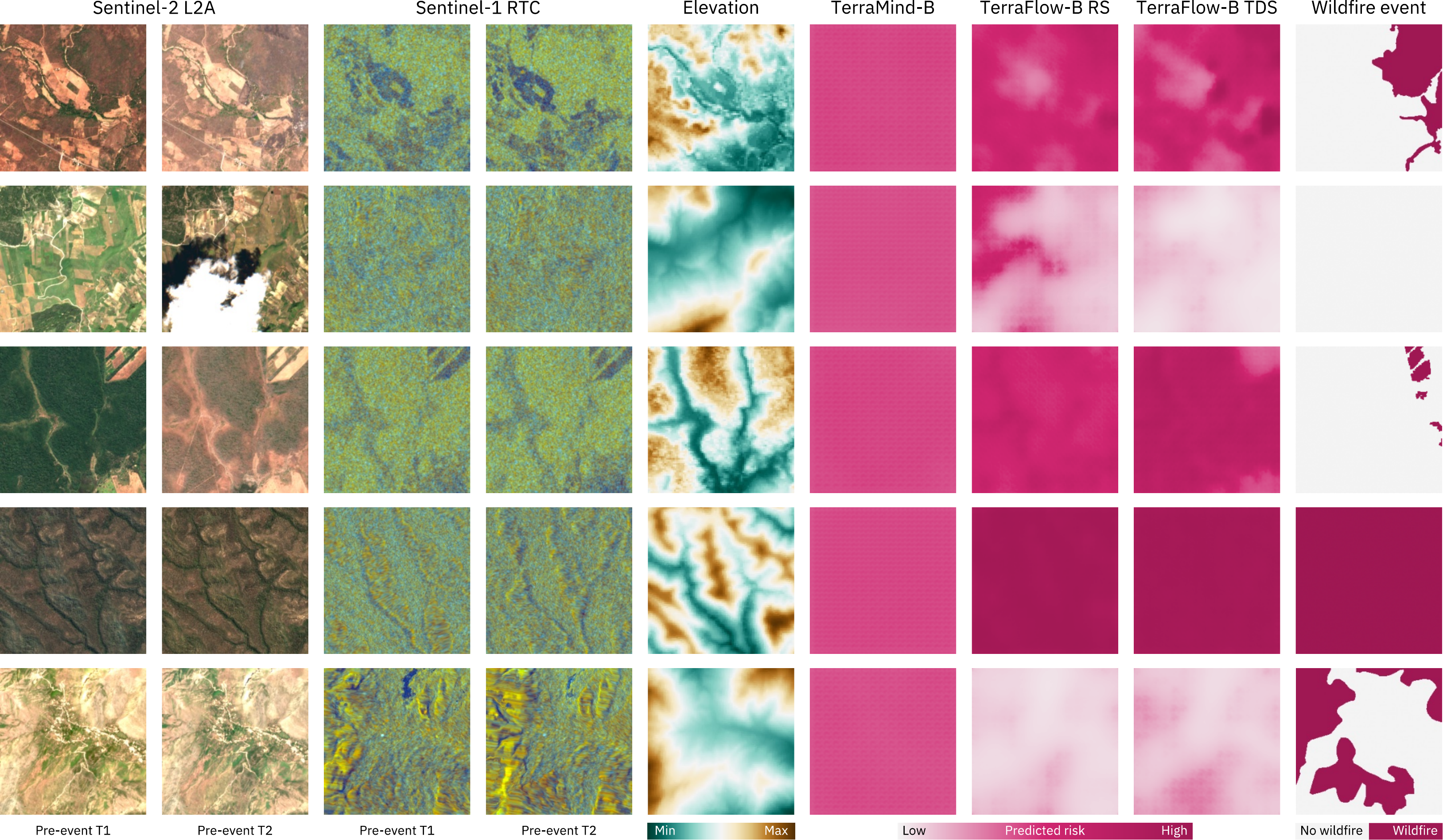}
    \caption{Qualitative comparison on the ImpactMesh-Fire risk prediction task. The inputs include two pre-event Sentinel-2 images, two Sentinel-1 images with VV-VH-VV/VH pseudo coloring, and DEM. 
    No model, including TerraFlow, is able to learn meaningful patterns for fire prediction which may stem from the unpredictable human influence on many fire events.}
    \label{fig:impactfires_comparison}
\end{figure}


\subsection{Temporal Analysis.}

\begin{figure}[!htb]
    \centering
    \begin{subfigure}{0.48\linewidth}
    \centering
    \includegraphics[width=\linewidth]{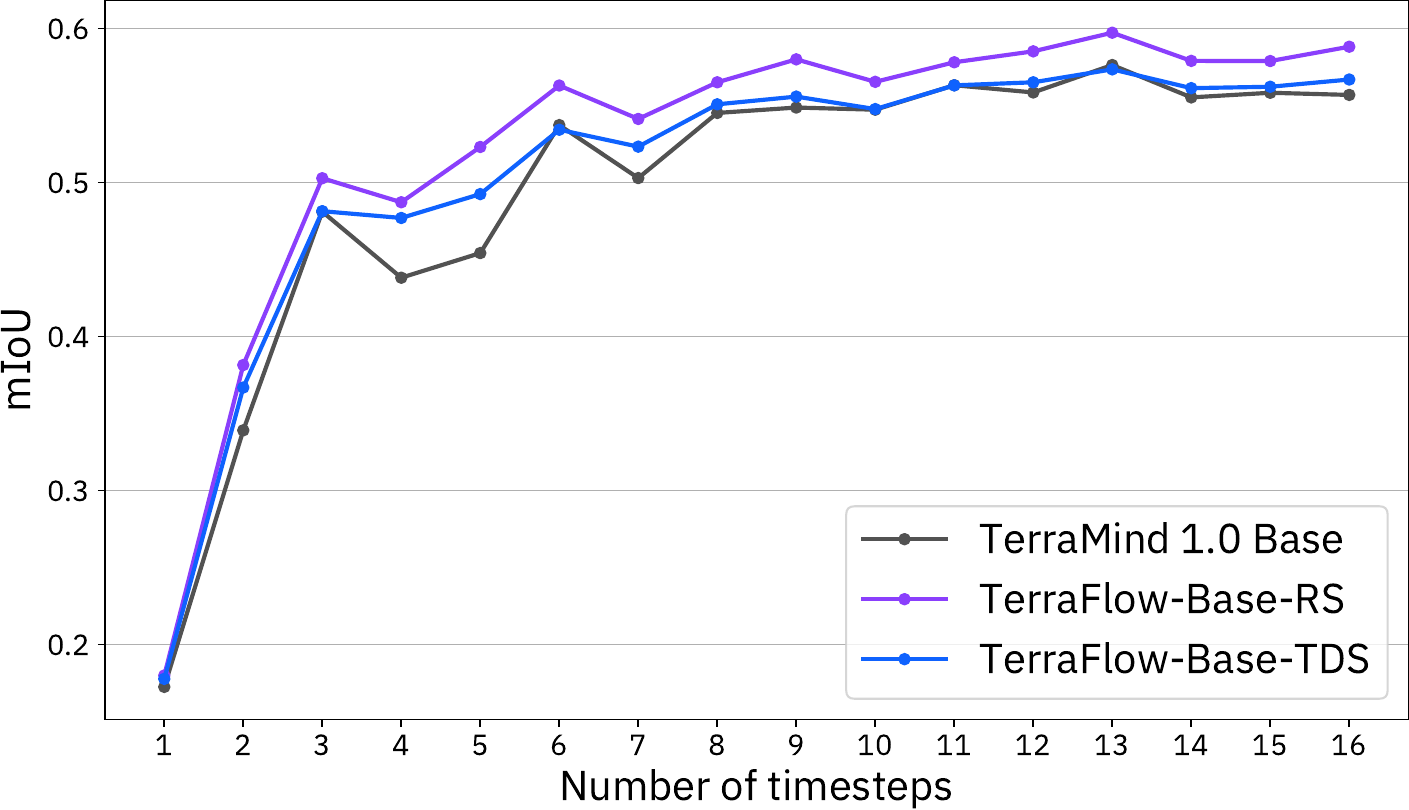}
    \caption{PASTIS}
    \label{subfig:pastis_timesteps}
    \end{subfigure}
  \hfill
    \begin{subfigure}{0.48\linewidth}
    \centering
    \includegraphics[width=\linewidth]{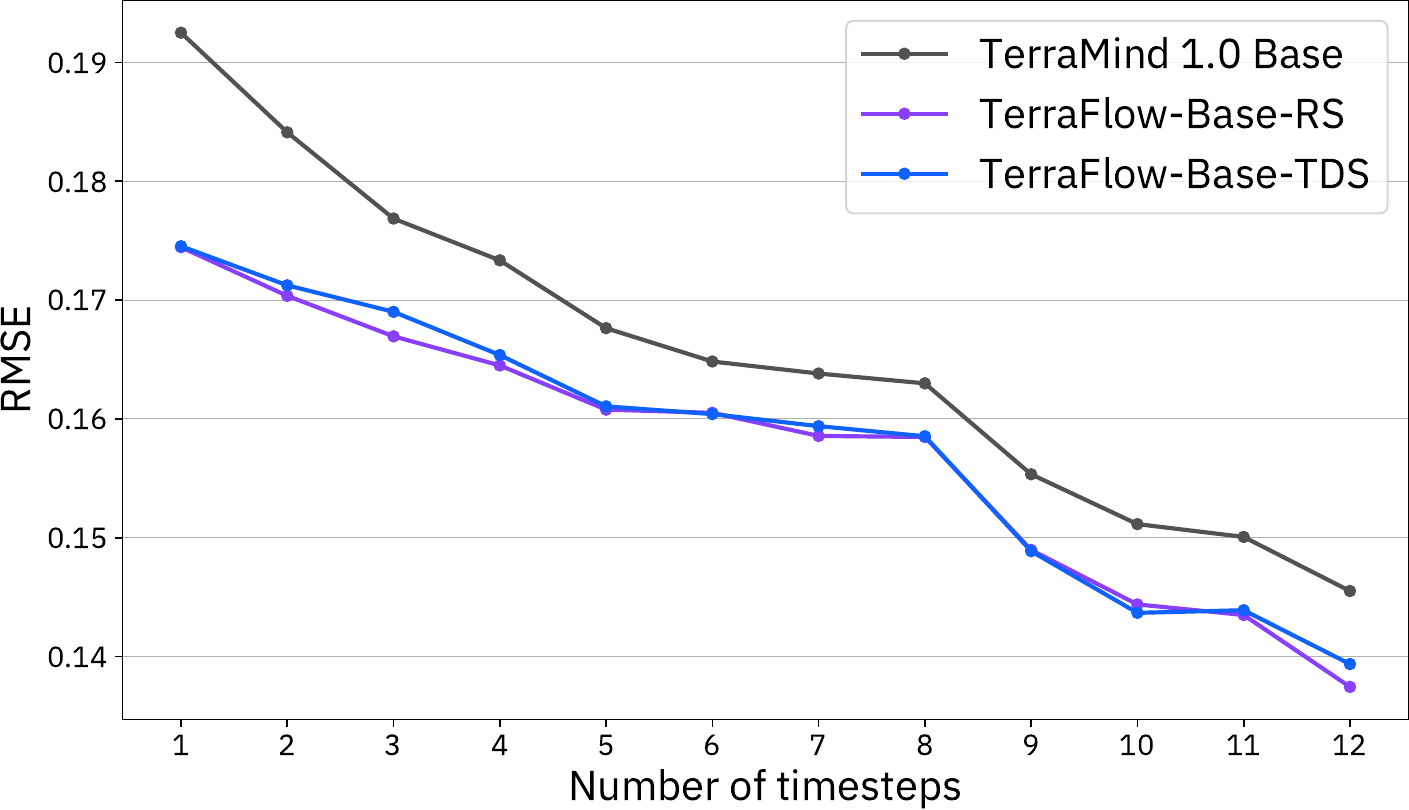}
    \caption{BioMassters}
    \label{subfig:bio_timesteps}
    \end{subfigure}
    \caption{Downstream task performance on test sets as a function of the number of timesteps per sample, uniformly sampled from the available set.}
    \label{fig:timesteps}
\end{figure}

\begin{figure}[!hbt]
    \centering
    \begin{subfigure}{0.48\linewidth}
    \centering
    \includegraphics[width=\linewidth]{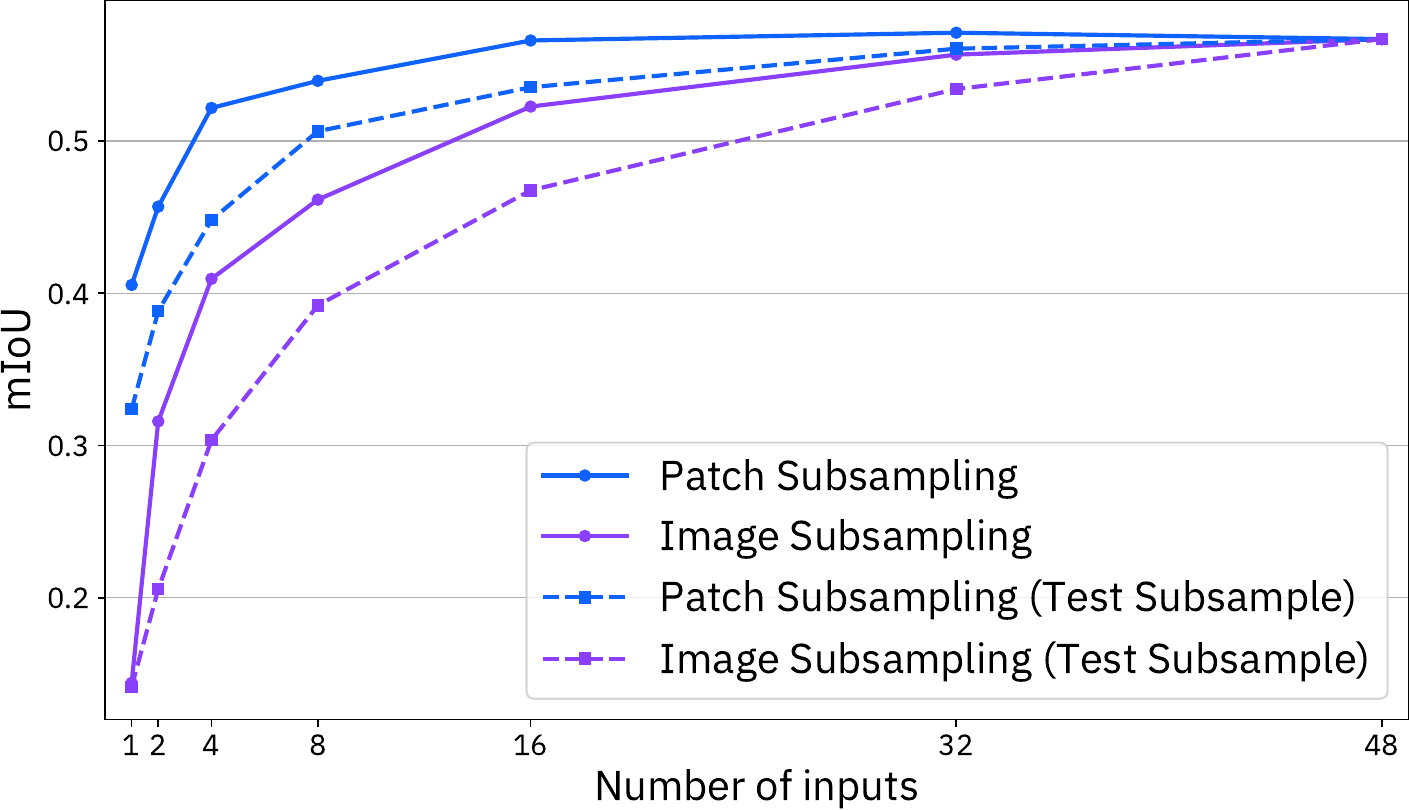}
    \caption{PASTIS}
    \label{subfig:pastis_subsampling}
    \end{subfigure}
  \hfill
    \begin{subfigure}{0.48\linewidth}
    \centering
    \includegraphics[width=\linewidth]{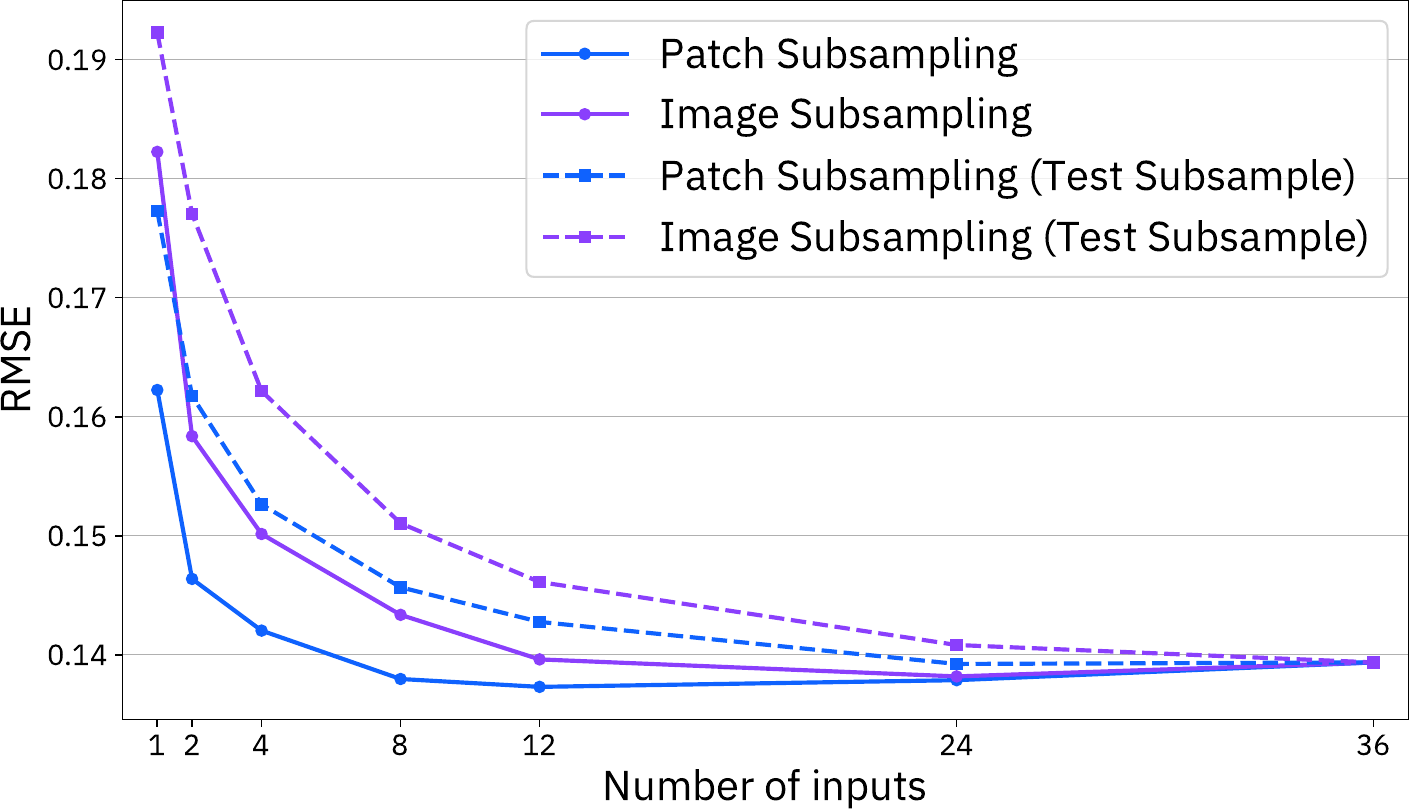}
    \caption{BioMassters}
    \label{subfig:bio_subsampling}
    \end{subfigure}
    \caption{Effect of token subsampling: Random image subsampling reduces training time at the cost of accuracy, patch-level subsampling across timesteps and modalities achieves better results with fewer tokens.}
    \label{fig:subsampling}
\end{figure}

We further explore how model performance changes as a function of the number of timesteps for BioMassters and PASTIS in Fig.~\ref{fig:timesteps}. We observe that, for both tasks, increasing the number of timesteps improves the prediction. Beyond $\approx$16 timesteps, the performance on PASTIS begins to saturate, whereas the Biomass task continues to improve in an approximately linear fashion. For figures, see  the supplementary material. The results suggest that the available amount of temporal context has (up to a task-specific saturation point) a strong impact on crop type prediction and biomass estimation. 
Note, though, a larger number of timesteps also increases the computational cost. In the following, we explore the effects of two approaches to reduce computation during fine-tuning.

Computational cost in transformer-based architectures is naturally connected to the input sequence length which quadratically impacts the attention calculations. Therefore, we note that token subsampling will directly reduce computational cost during fine-tuning. As the model decoder expects features across all spatial patches, we decide to subsample along the time and modality dimensions. 
In the proposed datasets, we have 16 and 12 timesteps and 3 modalities that we subsample across, resulting in 48 and 36 single images per sample.

As shown in Fig.~\ref{fig:subsampling}, subsampling entire images significantly reduces computation times for fine-tuning, and leads to an expected decrease in performance. To preserve as much performance as possible, we propose a randomized sampling strategy in which, for each spatial position, tokens are sampled randomly while preserving all spatial locations. This results in a fully random set of input tokens, from which the model learns the meaning of each token through space, time, and modality encodings. At test time, the full input is used. We observe that model performance can be maintained using only one third of the samples, which corresponds to roughly half the fine-tuning time per epoch. Furthermore, when subsampling is not applied at test time, performance further improves, indicating that although the model is trained on a subset of tokens (dashed lines), it is able to extrapolate the learned information to complete inputs (solid lines).

\section{Discussion and Concluding Remarks}
\label{sec:conclusion}

This work provides a path to integrate temporal pretraining into existing spatially-focused deep learning models for EO. By embedding temporal structure already during pretraining, TerraFlow consistently improves downstream performance on multitemporal tasks. Thus, the temporal objective reshapes how the encoder allocates capacity, improving sequence‑aware representation quality without increasing model size. Two design choices are central to these gains. First, early temporal fusion via temporal attention over sequences of cross‑modal tokens, which lets the model discover interactions across time and modality, outperforming late‑fusion heads that only merge information after per‑frame encoding. Second, Temporal Disjoint Sampling (TDS) prevents shortcut solutions that reconstruct targets using only information from the same timestamp, forcing the network to exploit temporal patterns (e.g., persistence, lags, and change patterns) rather than leaning on single‑timestep spatial correlations.






\bibliographystyle{splncs04}
\bibliography{references}

\clearpage
\clearpage
\setcounter{page}{1}

\title{Supplementary Material to TerraFlow}

\titlerunning{TerraFlow: Multimodal, Multitemporal Representation Learning for EO}


\author{Nazar Puriy\thanks{Equal contribution.}\inst{1,2} \and
Johannes Jakubik\repeatthanks\inst{1} \corr \and
Benedikt Blumenstiel\inst{1} \corr \and
Konrad Schindler\inst{2}
}

\authorrunning{N. Puriy, J. Jakubik, B. Blumenstiel, and K. Schindler}

\institute{IBM Research Europe, Zurich, Switzerland 
\and
ETH Zurich, Zurich, Switzerland
\email{\{johannes.jakubik1,benedikt.blumenstiel\}@ibm.com}
}

\maketitle

\appendix

\section{Masking Strategies}
\label{sec:masking_strategies}
During the pretraining phase, we explored different masking strategies and analyzed how they affect model performance. It is worth noting that changing the masking strategy alters the underlying task; therefore, final validation losses are not directly comparable across strategies. Due to computational constraints, we were unable to fully pretrain models with all masking strategies and evaluate them exhaustively on downstream tasks. Consequently, the selection of masking strategies was guided by logical considerations, complemented by short pretraining runs to assess their behavior. In particular, we examined learning dynamics (e.g., learning speed and convergence) and their impact on downstream performance.

\begin{figure}[!ht]
    \centering
    \includegraphics[width=\linewidth]{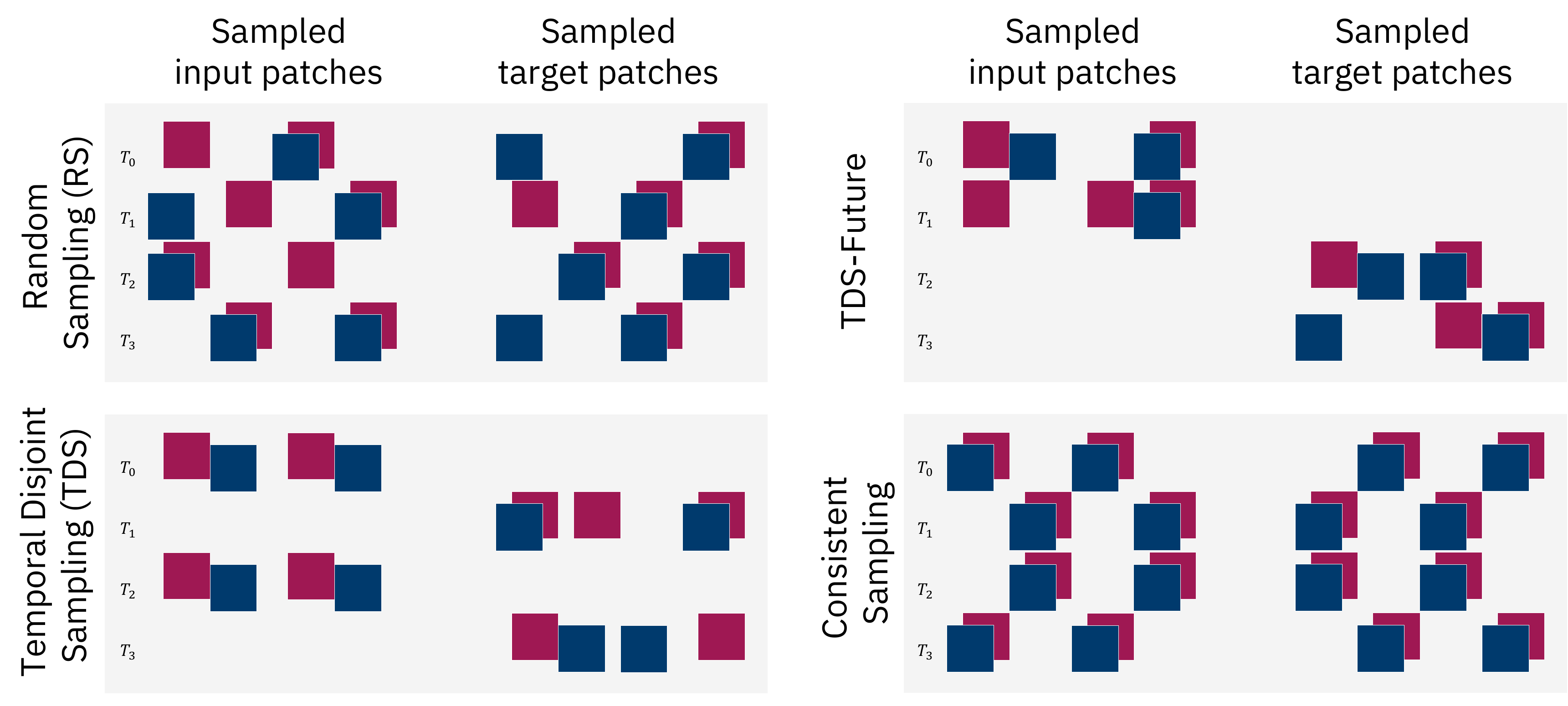}
    \caption{Strategies used during the pretraining phase. Random Sampling (RS), where tokens are randomly masked as inputs and outputs. Temporal Disjoint Sampling (TDS), where input and output tokens do not share timesteps. TDS-Future, where input tokens correspond to earlier timesteps than the target tokens. Consistent Sampling, where the same tokens are sampled across modalities.}
    \label{fig:sampling_strategies}
\end{figure}

Fig.~\ref{fig:sampling_strategies} illustrates the proposed masking strategies alongside two additional variants that were considered. The default and first strategy we evaluated is random sampling, where time is treated as an additional dimension and tokens are sampled uniformly across it. A potential drawback of this approach is that a pretrained model may rely primarily on already learned spatial and modal context for token reconstruction, effectively ignoring temporal information. To mitigate this issue, we introduced a second variant, termed TDS, in which the model is forced to use information from different timesteps when performing reconstruction.

Motivated by the disaster prediction task, which is inherently forward-looking, we also considered a variant similar to TDS but constrained such that input times always precede output times. However, this approach did not yield the expected improvements in preliminary experiments. We believe this is because the pretraining objective benefits from allowing the model to exploit the full temporal range when reconstructing masked tokens. Restricting sampling to strictly forward-looking configurations reduces the diversity of valid reconstruction contexts available during pretraining, without providing additional useful inductive bias, especially since the pretrained model is later used solely as a token embedding module, and additionally because all valid sampling configurations of this forward-only strategy are already encompassed by the base TDS formulation.

Finally, we explored a consistent cross-modality sampling strategy, where, when predicting a given token, the model is prevented from observing the corresponding tokens from other modalities. This approach resulted in worse performance in early experiments, likely because it caused the model to forget multimodal relationships too rapidly during pretraining.

\section{Experimental Setting: Downstream Applications}\label{appendix:exp_setting}

For downstream applications, we generally follow best practices and convert transformer features into a multi-scale spatial hierarchy using learned feature interpolation before decoding \cite{Simumba2025GeoBench2,Gomes2025TerraTorch}. From the 12-layer encoder, we extract four intermediate representations from layers $[2,5,8,11]$, capturing low-, mid-, and high level features. For semantic segmentation and regression benchmarking tasks, the extracted tokens are mean-pooled across modalities and time before being passed to the decoder following \cite{Simumba2025GeoBench2}. For disaster prediction, we employ feature concatenation rather than mean pooling in order to deal with the limited number of temporal samples available in this setting. 
Decoder channel widths are fixed to $[256,128,64,32]$ for TerraFlow-Tiny and $[512,256,128,64]$ for TerraFlow-Base. For semantic segmentation and regression tasks, we then attach a U-Net decoder~\cite{Ronneberger2015UNet}. 
The total model size is approximately 8M parameters for TerraFlow-Tiny and 100M parameters for TerraFlow-Base, with slight variations depending on the number of modalities used.
All downstream experiments are implemented using TerraTorch~\cite{Gomes2025TerraTorch}, which provides a unified interface for adapting geospatial models to GEO-Bench-2 tasks~\cite{Simumba2025GeoBench2}.\\

\noindent Training is performed on {Kuro Siwo} for 20 epochs and on {ImpactMesh} for 50 epochs, using early stopping with a patience of 10 epochs. Model selection is based on the {F1 score of the disaster class}. For each checkpoint, we additionally select the decision threshold that maximizes this F1 score on the validation set for later evaluation on the separate hold-out test set.
Beyond discrete classification performance, we also report the {Brier score}, which evaluates the quality of the predicted probabilities for the spatial risk maps. Unlike the F1 score, which depends on a fixed threshold and only assesses binary decisions, the Brier score measures the squared error between predicted probabilities and the true binary outcomes, thereby capturing both calibration and confidence of the model predictions. 
Formally, given predicted probabilities $p_i \in [0,1]$ and ground-truth labels $y_i \in \{0,1\}$ for $N$ spatial pixels, the Brier score is defined as $\mathrm{Brier} = \frac{1}{N} \sum_{i=1}^{N} \left(p_i - y_i \right)^2$.
This metric allows us to assess not only whether disasters are correctly detected, but also whether the predicted risk levels are meaningful and well-calibrated.

\section{Additional Results}
\label{app:aditional_results}

\begin{figure}[tbh]
    \centering
    \includegraphics[width=\linewidth]{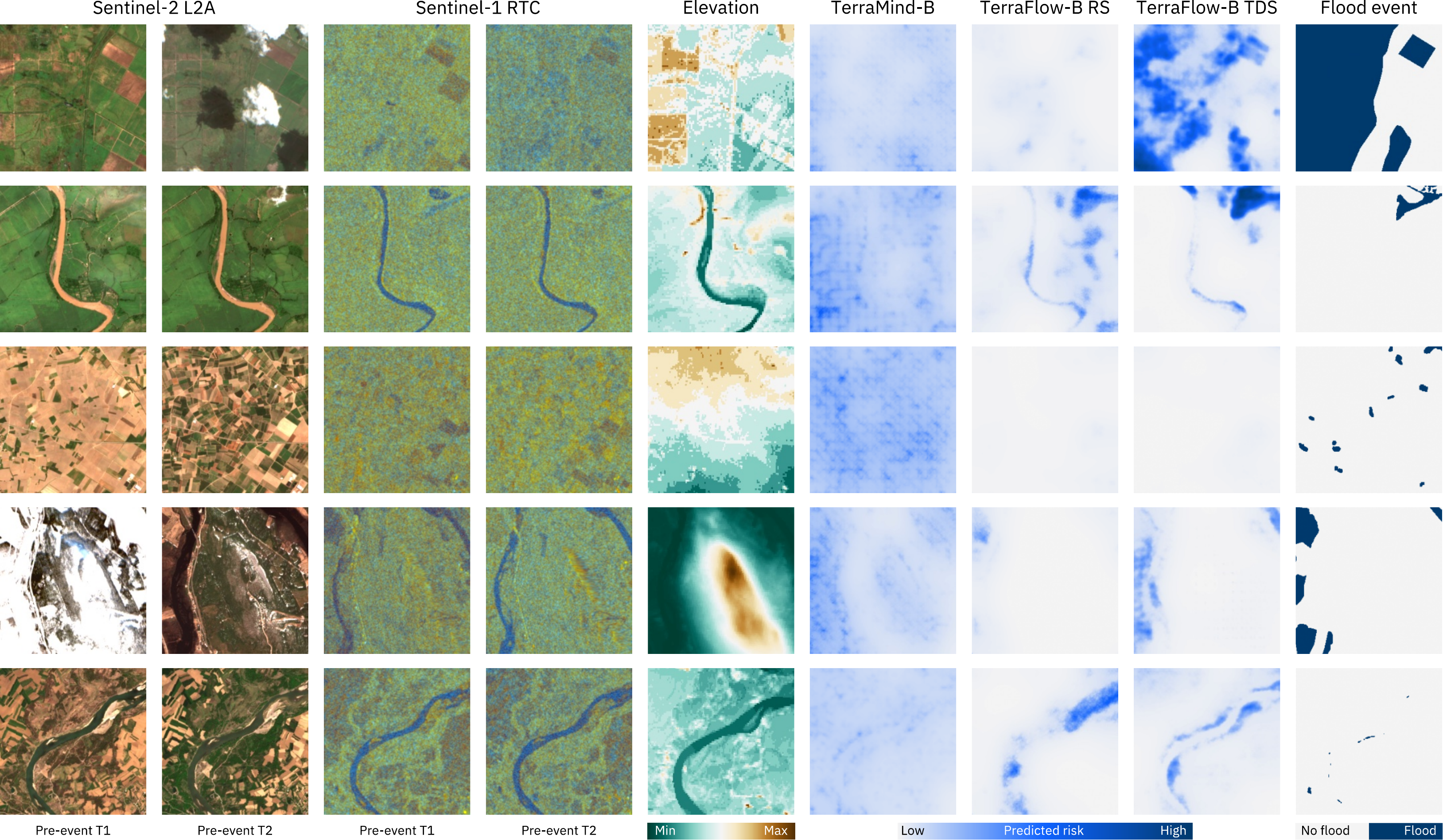}
    \caption{Qualitative comparison on the ImpactMesh-Flood risk prediction task. The inputs include two pre-event Sentinel-2 images, two Sentinel-1 images with VV-VH-VV/VH pseudo coloring, and DEM. We observe that the TerraFlow models are able to learn meaningful patterns for flood prediction while image-level models like TerraMind completely fail to accurately predict risk.}
    \label{fig:impactfloods_comparison}
\end{figure}

We provide qualitative examples of the ImpactMesh flood risk predictions in Fig.~\ref{fig:impactfloods_comparison}. The results are in line with the predictions on Kuro Siwo presented in the main paper. 
This section provides further insights from our experiments.

\subsection{Ablation Studies}

\noindent \textbf{Modality analysis.} \label{sec:modality}
Table~\ref{tab:modality_comparison} reports the performance obtained when varying the input modalities across both fire and flood prediction tasks. Overall, we observe that Sentinel-2 Level-2A is the most informative modality, consistently achieving the best performance in a unimodal setup. In some cases, combining S2 with additional modalities yields marginal improvements, although these gains are generally limited and not consistently significant. The lower performance of RTC can be explained by the fact that its weights were not trained for the pretraining task and correspond to the TerraMind checkpoint; therefore, its performance, and that of any combination containing it, is expected to be lower. We further observe that DEM alone yields the poorest performance, which can be attributed to overfitting on the training data. Using relative DEM, normalized per image, improves performance; however, the results remain insufficient, as DEM does not include information on the general presence of water in near proximity to low-altitude environments. However, interestingly, combining DEM with only one additional modality degrades performance, suggesting that elevation-related cues are already implicitly captured by the other modalities. In contrast, integrating all three modalities yields the best results, as they provide complementary information and, crucially, capture temporal dynamics that are absent in DEM alone.\\

\begin{table}[!ht]
\centering
\caption{TerraFlow-Base results on ImpactMesh flood and fire tasks show that multimodal inputs (S1+S2+DEM) perform best overall; Sentinel-2 is the strongest single modality, while DEM’s impact is task- and modality-dependent, benefiting floods mainly when combined with S2. We report F1$_{\{Flood,Fire\}}$~[\%]~$\uparrow$ as mean $\pm$ std across three runs.}
\small
\begin{tabularx}{\textwidth}{cccc CC|CC}
\toprule
\multirow{2}{*}{S1} &
\multirow{2}{*}{S2} &
\multirow{2}{*}{DEM} &
Rel. &
\multicolumn{2}{c}{ImpactMesh-Flood} &
\multicolumn{2}{c}{ImpactMesh-Fire} \\
 & & & DEM &
 TF-Base-RS & TF-Base-TDS & TF-Base-RS & TF-Base-TDS \\
\midrule

\ding{51} & \ding{55} & \ding{55} & \ding{55}
    & \cellcolor{Light!43!Dark}20.2 $\pm$ 0.3
    & \cellcolor{Light!47!Dark}21.3 $\pm$ 1.9
    & \cellcolor{Light!62!Dark}51.7 $\pm$ 2.3
    & \cellcolor{Light!69!Dark}52.5 $\pm$ 1.4 \\

\ding{55} & \ding{51} & \ding{55} & \ding{55}
    & \cellcolor{Light!96!Dark}38.2 $\pm$ 1.6
    & \cellcolor{Light!94!Dark}37.8 $\pm$ 2.9
    & \cellcolor{Light!88!Dark}54.7 $\pm$ 1.2
    & \cellcolor{Light!87!Dark}54.6 $\pm$ 3.0 \\

\ding{55} & \ding{55} & \ding{51} & \ding{55}
    & \cellcolor{Light!1!Dark}5.7 $\pm$ 1.3
    & \cellcolor{Light!0!Dark}5.3 $\pm$ 2.5
    & \cellcolor{Light!0!Dark}44.5 $\pm$ 3.4
    & \cellcolor{Light!18!Dark}46.6 $\pm$ 3.0 \\

\ding{55} & \ding{55} & \ding{55} & \ding{51}
    & \cellcolor{Light!36!Dark}17.8 $\pm$ 1.2
    & \cellcolor{Light!33!Dark}16.7 $\pm$ 0.9
    & \cellcolor{Light!38!Dark}49.0 $\pm$ 0.0
    & \cellcolor{Light!38!Dark}48.9 $\pm$ 0.1 \\

\ding{51} & \ding{51} & \ding{55} & \ding{55}
    & \cellcolor{Light!94!Dark}37.8 $\pm$ 1.2
    & \cellcolor{Light!100!Dark}39.7 $\pm$ 0.8
    & \cellcolor{Light!85!Dark}54.5 $\pm$ 3.4
    & \cellcolor{Light!94!Dark}55.4 $\pm$ 1.5 \\

\ding{51} & \ding{55} & \ding{51} & \ding{55}
    & \cellcolor{Light!15!Dark}10.5 $\pm$ 5.5
    & \cellcolor{Light!19!Dark}11.9 $\pm$ 1.8
    & \cellcolor{Light!53!Dark}50.7 $\pm$ 0.9
    & \cellcolor{Light!61!Dark}51.6 $\pm$ 1.8 \\

\ding{51} & \ding{55} & \ding{55} & \ding{51}
    & \cellcolor{Light!37!Dark}17.9 $\pm$ 2.6
    & \cellcolor{Light!32!Dark}16.4 $\pm$ 1.0
    & \cellcolor{Light!59!Dark}51.3 $\pm$ 0.9
    & \cellcolor{Light!13!Dark}46.1 $\pm$ 2.9 \\

\ding{55} & \ding{51} & \ding{51} & \ding{55}
    & \cellcolor{Light!98!Dark}38.9 $\pm$ 0.8
    & \cellcolor{Light!92!Dark}36.9 $\pm$ 2.1
    & \cellcolor{Light!82!Dark}54.0 $\pm$ 1.8
    & \cellcolor{Light!88!Dark}54.7 $\pm$ 0.6 \\

\ding{55} & \ding{51} & \ding{55} & \ding{51}
    & \cellcolor{Light!93!Dark}37.4 $\pm$ 0.8
    & \cellcolor{Light!94!Dark}37.8 $\pm$ 0.7
    & \cellcolor{Light!83!Dark}54.1 $\pm$ 0.8
    & \cellcolor{Light!90!Dark}54.9 $\pm$ 0.5 \\

\ding{51} & \ding{51} & \ding{51} & \ding{55}
    & \cellcolor{Light!100!Dark}39.7 $\pm$ 2.6
    & \cellcolor{Light!98!Dark}39.1 $\pm$ 2.2
    & \cellcolor{Light!91!Dark}55.0 $\pm$ 2.4
    & \cellcolor{Light!100!Dark}56.1 $\pm$ 0.9 \\

\ding{51} & \ding{51} & \ding{55} & \ding{51}
    & \cellcolor{Light!95!Dark}37.9 $\pm$ 1.1
    & \cellcolor{Light!96!Dark}38.2 $\pm$ 0.3
    & \cellcolor{Light!73!Dark}53.0 $\pm$ 1.8
    & \cellcolor{Light!93!Dark}55.3 $\pm$ 0.4 \\
\bottomrule
\end{tabularx}
\label{tab:modality_comparison}
\end{table}

\noindent \textbf{Shuffling across the temporal dimension.} We stress-test the temporal understanding of our models in order to understand whether the model is merely capturing temporal agreement or effectively models temporal dynamics. To investigate this, starting from the full set of images, we randomly shuffle timestamps, thereby providing the model with unordered image sets. Table~\ref{tab:temporal_comparison} shows that this shuffling generally causes only marginal drops in performance. Similar behavior has been observed in other temporal models~\cite{Szwarcman2024PrithviEO2}, indicating that the model possesses strong temporal understanding even without temporal positional encoding inferring temporal information directly from the input images themselves. We further note that the absence of performance decreases for BioMassters, while performance slightly declines from shuffling for PASTIS. This suggests that the model captures some form of temporal dynamics for PASTIS as crop-type segmentation inherently relies on the temporal evolution of vegetation, whereas for BioMassters it mainly relies on temporal agreement where model performance depends on aggregated structural and spectral information.

Overall, we interpret the findings from the ablation experiments as a certain temporal robustness of TerraFlow. Although the model is trained on sequences of length four, it is able to extrapolate to sequences of arbitrary length. Empirically, the model (1)~consistently outperforms competing approaches across tasks involving variable sequence lengths, and (2)~exhibits no detectable bias toward processing sequences of length four, nor any pronounced bias with respect to the temporal spacing between consecutive samples. 


\begin{table}[t]
\centering
\setlength{\tabcolsep}{4pt}
\caption{Comparison with TerraFlow-Base-TDS between the full, subsampled, and timestamp-shuffled variants using five timesteps on PASTIS and BioMassters. Performance is comparable overall, with subsampling yielding similar or slightly better results, while timestamp shuffling highlights differing sensitivity to temporal order across tasks.}
\label{tab:temporal_comparison}
\begin{tabular}{lcc}
\toprule
 & PASTIS & BioMassters \\
Input & mIoU [\%] ↑ & RMSE ↓ \\
\midrule
Five timesteps & 56.5 $\pm$ 0.1 & 0.138 $\pm$ 0.000 \\
\makecell{$\frac{1}{3}$ Patch Subsampling} & 56.6 $\pm$ 0.3 & 0.137 $\pm$ 0.000 \\
Time-shuffled images & 55.7 $\pm$ 0.1 & 0.138 $\pm$ 0.000 \\
\bottomrule
\end{tabular}
\end{table}

\subsection{Preliminary experiments}
One of the initial concerns of this work was whether pretraining a temporal version of the model would be beneficial. This concern arose from the fact that TerraMind 1.0 Base already outperformed all other models on GEO-Bench-2, making it unclear whether temporal pretraining would yield additional improvements or whether simple early-time fusion would be sufficient. Specifically, we aimed to determine whether temporal pretraining would improve final performance, accelerate convergence, or provide no measurable benefit.

\begin{figure}[!ht]
    \centering
    \includegraphics[width=0.8\linewidth]{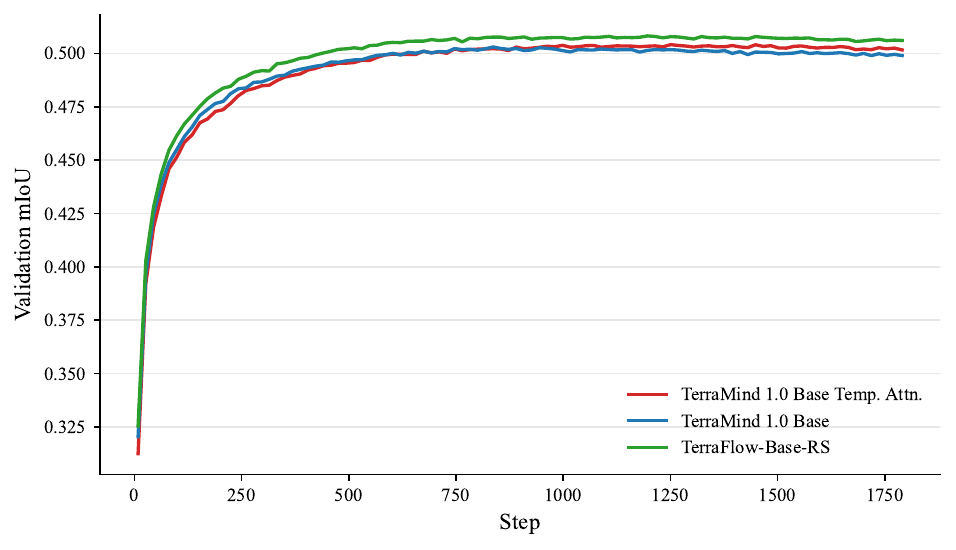}
    \caption{Preliminary MultiCrop results comparing TerraMind 1.0 Base, TerraMind with temporal attention, and TerraFlow-Base-RS, showing slower early convergence with temporal attention but improved final performance, and a larger gain when initializing from pretrained TerraFlow-Base weights.}
    \label{fig:multicrop}
\end{figure}

To address this question, we conducted preliminary experiments on the MultiCrop dataset~\cite{hls-multi-temporal-crop-classification}, a multitemporal crop classification benchmark based on Harmonized Landsat–Sentinel imagery over the contiguous United States in 2022. Similarly to \cite{Szwarcman2024PrithviEO2}, we train the models for 80 epochs using a batch size of 8 and a learning rate of 0.0001. Fig.~\ref{fig:multicrop} compares TerraMind 1.0 Base, TerraMind augmented with temporal attention, and TerraFlow-Base-RS. We observe, Table~\ref{tab:early_comparison}, that incorporating temporal attention slows early convergence, as additional temporal dependencies must be learned, but ultimately improves performance from 50.3 to 50.4. In contrast, initializing from pretrained TerraFlow-Base weights leads to a larger and more substantial improvement, achieving a score of 50.8.

\begin{table}[h]
\centering
\setlength{\tabcolsep}{4pt}
\caption{Early comparison of segmentation performance on the MultiCrop dataset. Results for UNet, Prithvi, TerraMind 1.0 Base (with and without temporal attention), and TerraFlow-Base-RS. The results show that introducing early temporal attention~(TA) combined with pretraining improves performance, with TerraFlow-Base-RS achieving performance comparable to Prithvi-EO-2.0-600M.}
\label{tab:early_comparison}
\begin{tabular}{lcc}
\toprule
Model & TA & mIoU $\uparrow$ \\
\midrule
UNet & \ding{55} & 42.6 \\
Prithvi-EO-1.0-100M & \ding{51} & 42.7 \\
Prithvi-EO-2.0-300M & \ding{51} & 48.6 \\
Prithvi-EO-2.0-600M & \ding{51} & 50.7 \\
\midrule
TerraMind 1.0 Base & \ding{55} & 50.3 $\pm$ 0.04 \\
TerraMind 1.0 Base & \ding{51} & 50.4 $\pm$ 0.08 \\
TerraFlow-Base-RS & \ding{51} & 50.8 $\pm$ 0.01 \\
\bottomrule
\end{tabular}
\end{table}

\subsection{Static Disaster Prediction}

In Table~\ref{tab:1timestamps_results}, we show a comparison between using two pre-event images and using only one. We observe that using only a single image significantly decreases model performance. However, we note that part of the performance drop may come from the fact that some input samples are cloudy, and when only one pre-event image is available, other images cannot be used to recover missing information. Although this is the case, given the results in Table~\ref{tab:impactmesh_comparison}, we would also expect part of the performance drop to stem from the fact that temporal dynamics are not used to predict the events. However, we cannot draw a definitive conclusion from these results alone.

\begin{table}[!ht]
\centering
\setlength{\tabcolsep}{4pt}
\caption{Model comparison for TerraFlow-Base-TDS using 1 or 2 pre-event timestamps for flood and fire disaster prediction tasks.}
\label{tab:1timestamps_results}
\begin{tabular}{lcc}
\toprule
Input & Floods mIoU ↑ & Fire mIoU ↑ \\
\midrule
One pre-event timestep & 34.9 ± 0.6 & 53.4 ± 0.7 \\
Two pre-event timesteps & \textbf{39.7 ± 0.4} & \textbf{56.6 ± 0.3} \\
\bottomrule
\end{tabular}
\end{table}

\subsection{Kuro Siwo Segmentation}\label{appendix:kuro_siwo_segmentation}
We furthermore evaluate our approach on the Kuro Siwo  \textbf{flood segmentation} task, where the objective is to predict three semantic classes—no water (NW), permanent water (PW), and flood (F)—from pairs of pre- and \textbf{post-event} SAR images. We compare our model against the best-performing architectures reported in BlackBench \cite{Bountos2024KuroSiwo}, which provides an extensive and diverse benchmark for this dataset.\\

\noindent BlackBench includes a wide range of architectures, spanning standard semantic segmentation models such as U-Net, DeepLabv3, and UPerNet with both convolutional and transformer-based backbones, as well as models explicitly designed for change detection, including FC-EF variants, SNUNet-CD, Changeformer, and recurrent architectures such as ConvLSTM. While standard segmentation models typically exploit temporal information implicitly through channel stacking, change detection and recurrent models introduce explicit temporal learning via dual-branch encoders or sequential processing. Given the breadth and maturity of this benchmark, Kuro Siwo can be considered close to saturation, motivating multiple training runs of our models to account for stochastic effects and ensure robust performance estimates.\\

\noindent As shown in Table~\ref{tab:appendix_kurosiwo_comparison}, TerraFlow with temporal attention (TerraFlow-Base-TDS) achieves the best overall performance, surpassing the strongest BlackBench baseline (UNet-ResNet50) in terms of mIoU and F1-F, while remaining competitive across all other metrics. This highlights the importance of explicitly modeling temporal dependencies between pre- and post-event SAR acquisitions for flood segmentation. We again observe a similar learning behavior to that of the flood prediction task, in which TerraFlow achieves the best performance, TerraFlow with random weights performs the worst, and TerraFlow initialized with TerraMind weights achieves intermediate performance.

\begin{table}[tbh]
    \centering
    \caption{
    Comparison of the Kuro Siwo segmentation task against the best model results reported in the original paper~\cite{Bountos2024KuroSiwo}, together with TerraFlow and TerraMind variants.
    Best values are shown in \textbf{bold}, second-best values are \underline{underlined}.}
    \label{tab:appendix_kurosiwo_comparison}
    \begin{tabularx}{\textwidth}{lccCCCC}
        \toprule
        Model & PT & TA & F1$_{NW}$ [\%] ↑ & F1$_{PW}$ [\%] ↑ & F1$_{Flood}$ [\%] ↑ & mIoU [\%] ↑ \\
        \midrule
        UNet-ResNet18 & \ding{51} & \ding{55} & 98.7 & 76.0 & 79.9 & 75.1 \\
        UNet-ResNet50 & \ding{51} & \ding{55} & 98.7 & \underline{78.2} & \underline{80.1} & \underline{76.2} \\
        UNet-ResNet101 & \ding{51} & \ding{55} & 98.7 & \textbf{79.3} & 78.9 & 76.1 \\
        DeepLab-ResNet18 & \ding{51} & \ding{55} & \underline{98.7} & 77.4 & 78.5 & 75.1 \\
        \midrule
        ViT-Base & \ding{55} & \ding{51} & 98.6 ± 0.1 & 74.9 ± 0.5 & 76.6 ± 0.9 & 73.1 ± 0.5 \\
        TerraMind 1.0 Base & \ding{51} & \ding{55} & 98.6 ± 0.1 & 75.7 ± 1.2 & 77.4 ± 1.1 & 73.8 ± 0.9 \\
        TerraMind 1.0 Base & \ding{51} & \ding{51} & 98.6 ± 0.1 & 75.0 ± 1.1 & 78.4 ± 1.3 & 74.3 ± 1.2 \\
        TerraFlow-Base-RS & \ding{51} & \ding{51} & 98.7 ± 0.1 & 76.0 ± 1.5 & 80.0 ± 0.3 & 75.1 ± 0.8 \\
        TerraFlow-Base-TDS & \ding{51} & \ding{51} & \textbf{98.8 ± 0.0} & 77.9 ± 0.5 & \textbf{81.9 ± 0.6} & \textbf{77.0 ± 0.3} \\
        \bottomrule
    \end{tabularx}
\end{table}

\section{Analysis of GEO-Bench-2 Performance Differences}
\label{app:geobench2}
Because we obtained statistically significant differences for TerraMind 1.0 Base on the DynamicEarthNet and PASTIS tasks, we conducted additional experiments and analysis. For all experiments, we used the officially provided data-loading modules and dataset splits. We contacted the original paper's authors and obtained further information regarding their experimental setup. We observed that, for TerraMind models on DynamicEarthNet, the full set of input data was not used: the Planet Lab data were mapped into the Sentinel-2 L2A modality, thereby entirely omitting the Sentinel-2 L2A input module. For PASTIS, one identified factor was the sampling strategy: the authors' initial strategy sampled the most recent images rather than sampling uniformly. We re-ran the experiments under these two assumptions and still obtained significantly better results, as shown in Table~\ref{tab:geobench_comparison_v2}. We therefore believe that additional, unaccounted-for factors remain, and for this reason we do not include a full quantitative comparison for these two datasets in the main results table, as we consider the originally reported results to be unreliable and potentially affecting the performance estimates of other models as well. Nevertheless, we note that on the publicly available GEO-Bench-2 leaderboard, TerraMind models appear among the best-performing approaches; accordingly, we would expect performance trends to scale in a broadly similar manner across evaluations. Accordingly, our models again would outperform those evaluated on these tasks.

\begin{table}
\centering
\setlength{\tabcolsep}{4pt}
\caption{Comparison between our originally reported TerraMind results in mIoU and a second configuration called Rerun, together with the reported results on GEO-Bench-2, where TerraMind 1.0 Large is the best-performing model inside the GEO-Bench-2 benchmark.}
\label{tab:geobench_comparison_v2}
\begin{tabular}{lcc}
\toprule
Model & PASTIS & DEN \\
\midrule
TerraMind 1.0 Base (GEO-Bench-2) & $42.2 \pm 0.3$ & $23.3 \pm 2.2$ \\
TerraMind 1.0 Large (GEO-Bench-2) &  $43.1 \pm 0.1$ & $34.5 \pm 0.6$ \\
\midrule
TerraMind 1.0 Base & $54.3 \pm 0.5$ & $35.8 \pm 1.1$ \\
TerraMind 1.0 Base-Rerun & $54.5 \pm 0.3$ & $32.0 \pm 1.4$ \\
\bottomrule
\end{tabular}
\end{table}

We also provide in Table~\ref{tab:biomassters_raw} the raw BioMassters RMSE values, which show in greater detail the differences between models that are not visible when using the aggregated metrics reported in Table~\ref{tab:geobench_results}. We observe a larger difference between TerraMind 1.0 Tiny and TerraFlow-Tiny, and that TerraFlow-Tiny-TDS outperforms TerraFlow-Tiny-RS and both outperform TerraMind 1.0 Base.

\begin{table}
\centering
\caption{Raw BioMassters results reporting RMSE (Mean $\pm$ Std)}
\label{tab:biomassters_raw}
\begin{tabular}{lc}
\toprule
Model & RMSE$\downarrow$ \\
\midrule
TerraMind 1.0 Tiny & $16.7 \pm 0.0$ \\
TerraFlow-Tiny-RS & $16.2 \pm 0.0$ \\
TerraFlow-Tiny-TDS & $16.1 \pm 0.0$ \\
TerraMind 1.0 Base & $16.3 \pm 0.0$ \\
TerraFlow-Base-RS & \boldmath{$15.8 \pm 0.0$} \\
TerraFlow-Base-TDS & \boldmath{$15.8 \pm 0.0$} \\
\bottomrule
\end{tabular}
\end{table}

\section{Runtime}
\begin{figure}[tbh]
    \centering
    \includegraphics[width=0.8\linewidth]{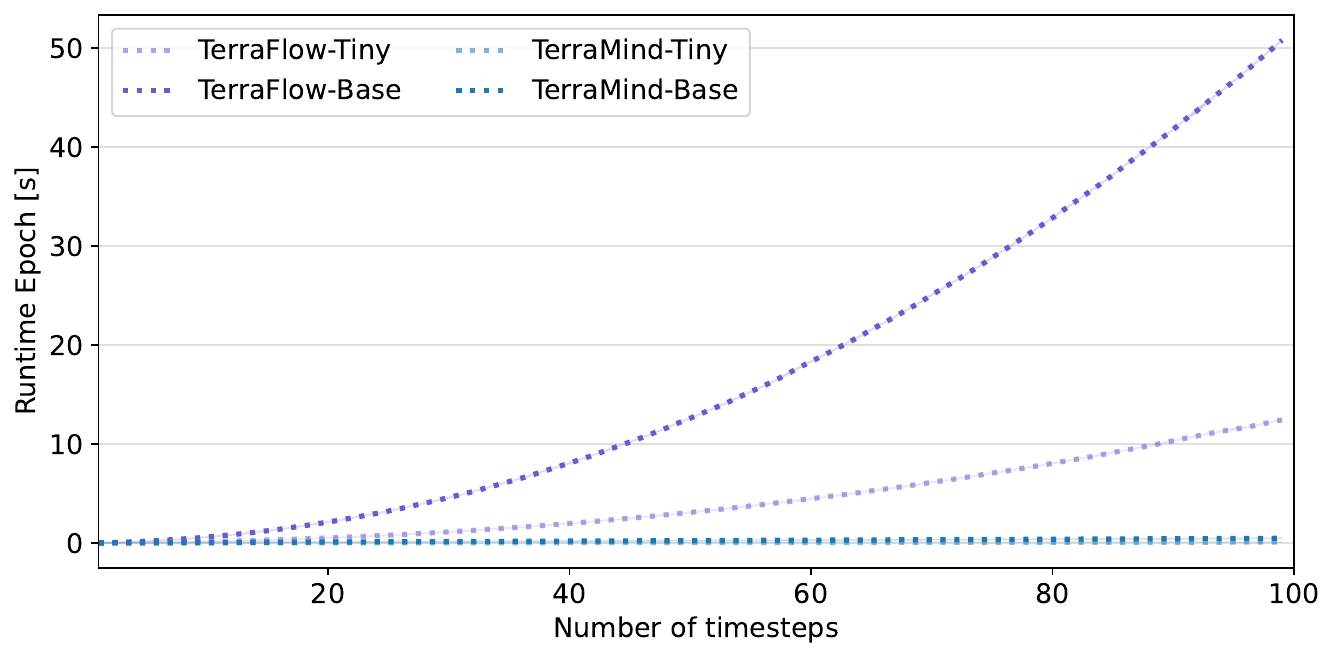}
    \caption{Mean runtime per training epoch in inference mode as a function of the number of timesteps 
$T$. Results are averaged over 50 runs with batch size 8. TerraMind exhibits approximately linear scaling since attention is not applied along the temporal dimension, whereas TerraFlow shows quadratic runtime growth due to temporal attention.}
    \label{fig:run_times}
\end{figure}

In Fig.~\ref{fig:run_times}, we report the inference time per epoch in no-gradient mode for batches of size 8 on a multimodal mock dataset containing S2L2A and SEN1GRD samples, as a function of the number of timesteps. The results show that the model execution time grows quadratically with the temporal length, which motivates the exploration of more efficient training strategies.

\vspace*{1.5cm}

\section{Further Remarks and Outlook}

On the temporal tasks of GEO‑Bench‑2, TerraFlow improves over multimodal, multi‑temporal, and single‑modality baselines. We further explore spatial risk‑map prediction as a stress test beyond conventional segmentation and regression tasks. Although these forecasting‑like tasks are intentionally simplified and not meant as a substitute for operational systems, TerraFlow provides an initial step towards deep learning baselines for forward‑looking spatial risk estimation from pre‑event context (e.g., flood risk), demonstrating that temporal generative pretraining extends to anticipatory tasks where native temporal understanding is instrumental.

A second theme in our results is parameter efficiency. TerraFlow‑Tiny approaches or exceeds TerraMind‑Base on multiple temporal benchmarks, indicating that temporal structure can partially substitute for model capacity. This suggests a favorable efficiency frontier for models that incorporate temporal objectives early in training, especially appealing for practitioners working under compute constraints. We also observe temporal robustness. Despite being continually pretrained on sequences of length four, TerraFlow generalizes well to variable‑length and irregularly sampled sequences at fine-tuning and test time. This is crucial in EO, where observation cadence is largely determined by revisit schedules and cloud cover. On top of that, we observe a robustness of TerraFlow against shuffling across the temporal dimension. This indicates that TerraFlow captures temporal agreement and, where necessary (e.g., crop phenology), largely maintains temporal ordering signals.
Finally, we show that fine-tuning efficiency can be retained. Token‑level temporal subsampling preserves most of the downstream accuracy while significantly reducing fine-tuning cost, making TerraFlow more practical in settings with limited resources. Together, these observations indicate that explicit temporal pretraining, enforced by TDS and enabled by early fusion, yields better sequence representations, improved downstream results, and competitive computational costs in fine-tuning.

A promising avenue for future work lies in expanding the scale and scope of temporal pretraining itself. In particular, TerraFlow would benefit from training on substantially denser and longer temporal archives, allowing the model to capture richer seasonal, interannual, and long‑term dynamics. Such an extension would also enable the framework to support long‑horizon applications such as land‑use transitions, vegetation cycles, and climate‑relevant environmental monitoring. Beyond increasing the scale of temporal data in pretraining, we see an opportunity to strengthen the temporal learning objective, either by extending Temporal Disjoint Sampling or by designing new objectives that more explicitly capture changes or approach causal temporal structure. For actual disaster‑oriented applications, TerraFlow needs to be enhanced by hybrid modeling approaches that integrate data‑driven sequence representations with physical priors or simulation‑based components, in order to move from the current exploratory risk‑mapping experiments beyond deep‑learning toy examples. Finally, future work may explore temporal tokenizers or temporally conditioned generative decoders, which could provide more fine‑grained temporal abstractions and enable flexible forecasting or counterfactual analysis.\\

\noindent\textbf{Use of Large Language Models}

\noindent We used an LLM solely for grammar/proofreading to improve readability; all scientific content is original and the authors take full responsibility for it.

\end{document}